%% file: acl_latex.tex
\pdfoutput=1

\documentclass[11pt]{article}

\usepackage[final]{acl}

\usepackage{times}
\usepackage{latexsym}
\usepackage{amsmath}
\usepackage{enumitem}
\usepackage{amssymb} 
\usepackage{mathrsfs}
\usepackage{booktabs}
\usepackage{multirow} 
\usepackage{colortbl}
\usepackage{xcolor}
\usepackage{makecell}  
\usepackage{bbding}  
\usepackage{bbm}
\usepackage[ruled,linesnumbered]{algorithm2e}

\usepackage[T1]{fontenc}

\usepackage[utf8]{inputenc}

\usepackage{microtype}

\usepackage{inconsolata}

\usepackage{graphicx}
\definecolor{blue}{HTML}{074799}   
\definecolor{red}{HTML}{C62E2E}   
\definecolor{lightblue}{HTML}{C4D9FF}  
\definecolor{lightred}{HTML}{FAE0DF}

\title{
Benchmarking Multimodal RAG through a Chart-based Document Question-Answering Generation Framework
}

\author{Yuming Yang\textsuperscript{1}, Jiang Zhong\textsuperscript{1}\thanks{Corresponding author}, \textbf{Li Jin}\textsuperscript{2*}, \textbf{Jingwang Huang}\textsuperscript{1}, Jingpeng Gao\textsuperscript{1}\\
\textbf{Qing Liu\textsuperscript{2}}, \textbf{Yang Bai\textsuperscript{2}}, \textbf{Jingyuan Zhang\textsuperscript{3}}, \textbf{Rui Jiang\textsuperscript{1}}, \textbf{Kaiwen Wei}\textsuperscript{1*}  \\
\textsuperscript{1} College of Computer Science, Chongqing University, China \\
\textsuperscript{2} Aerospace Information Research Institute, Chinese Academy of Sciences, China \\
\textsuperscript{3} Kuaishou Technology, Beijing, China \\
\text{\href{mailto:ymyang@cqu.edu.cn}{ymyang@cqu.edu.cn}, \href{mailto:zhongjiang@cqu.edu.cn}{zhongjiang@cqu.edu.cn}, \href{mailto:weikaiwen@cqu.edu.cn}{weikaiwen@cqu.edu.cn}} \\
\text{\href{mailto:jinlimails@gmail.com}{jinlimails@gmail.com}} \\
}

\begin{document}
\maketitle
\begin{abstract}
Multimodal Retrieval-Augmented Generation (MRAG) enhances reasoning capabilities by integrating external knowledge. However, existing benchmarks primarily focus on simple image-text interactions, overlooking complex visual formats like charts that are prevalent in real-world applications. In this work, we introduce a novel task, \textbf{Chart-based MRAG}, to address this limitation. 
To semi-automatically generate high-quality evaluation samples, we propose \textbf{CHAR}t-based document question-answering \textbf{GE}neration (CHARGE), a framework that produces evaluation data  through structured keypoint extraction, crossmodal verification, and keypoint-based generation. 
By combining CHARGE with expert validation, we construct \textbf{Chart-MRAG Bench}, a comprehensive benchmark for chart-based MRAG evaluation, featuring 4,738 question-answering pairs across 8 domains from real-world documents.
Our evaluation reveals three critical limitations in current approaches: (1) unified multimodal embedding retrieval methods struggles in chart-based scenarios, (2) even with ground-truth retrieval, state-of-the-art MLLMs achieve only 58.19\% Correctness and 73.87\% Coverage scores, and (3) MLLMs demonstrate consistent text-over-visual modality bias during Chart-based MRAG reasoning. 
The CHARGE and Chart-MRAG Bench are released at \url{https://github.com/Nomothings/CHARGE.git}.

\end{abstract}

\section{Introduction}

Multimodal retrieval-augmented generation (MRAG) \cite{zhao2023retrieving} enhances multimodal reasoning by retrieving relevant external knowledge, and leveraging multimodal large language models (MLLMs) for informed response generation~\cite{openai2023gpt,zhang2024mm}. This approach substantially mitigates hallucinations and improves factual grounding~\cite{gao2023retrieval}. 

\begin{figure}[t]
\centering
  \includegraphics[width=\linewidth]{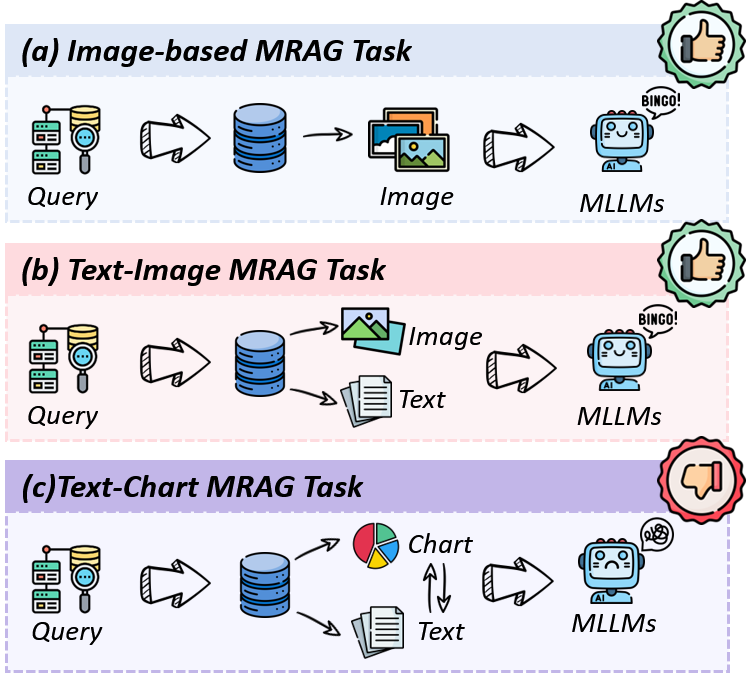}
  \caption{
    Comparison of two common MRAG scenarios, image-only and text-image, and the proposed text-chart task. In the text-chart MRAG scenario, models need to capture intricate chart details and retrieve both chart and text information to generate correct answers.
  }
  \label{fig:intro}
\end{figure}

Effectively evaluating MRAG systems requires high-quality benchmarks that assess both retrieval and generation. 
Existing benchmarks such as MRAG-Bench \cite{hu2024mragbench} and Dyn-VQA \cite{li2024benchmarking} have made strides in assessing MRAG capabilities through manually curated question-answering (QA) pairs. 
However, as illustrated in Fig.\ref{fig:intro}(a) and (b), these benchmarks primarily focus on scenarios involving images or simple combinations of images and text. Such settings fail to capture the complex interactions between visual details and corresponding text, particularly when dealing dense and structured information like charts, which are widely used in real-world applications \cite{masry2022chartqa}. This leaves a critical gap in MRAG evaluation. 

To bridge this gap, we propose a new task: \textbf{Chart-based MRAG}. For a given text query, this task involves three RAG sub-tasks: (1) \textit{Text-Chart MRAG}, as illustrated in~Fig.~\ref{fig:intro}(c), both textual and chart data must be jointly retrieved to generate correct answers. In addition, to allow for the separate evaluation of each modality's contributions, it also provides (2) \textit{Text-only RAG}, where answers can only be found in textual information; and (3) \textit{Chart-only MRAG}, where answers depend  exclusively on chart data. 
To comprehensively evaluate these tasks, a major challenge is how to semi-automatically generate high-quality QA pairs that accurately capture text-chart interactions.

To overcome this challenge, we propose \textbf{CHAR}t-based document question-answering \textbf{GE}neration (CHARGE), a framework for automatically generating QA pairs from real-world chart-document data. CHARGE follows a three-stage pipeline comprising structured keypoint extraction from text and chart data, crossmodal verification for accuracy, and keypoint-based generation to model complex multimodal interactions. Moreover, to further challenge the chart-based MRAG task, MLLMs are employed to generate QA pairs that require multi-hop reasoning based on intra-document or inter-document retrieval.

Building on CHARGE, we introduce \textbf{Chart-MRAG Bench}, a high-quality, human-checked benchmark tailored for Chart-based MRAG. With CHARGE, 5,866 qualified QA pairs were initially generated, after that, 4,738 (nearly 80\%) were meticulously selected through expert evaluation based on clarity, accuracy, multimodal coherence, and ethical considerations. As shown in Table~\ref{tab:data_compare}, Chart-MRAG Bench comprises 267 documents spanning 8 domains, 8 types of questions, 1,283 paragraphs, and 627 charts, capturing complex crossmodal interactions in realistic scenarios.

\input{table_dataset}
We conducted a systematic evaluation of mainstream retrieval methods and MLLMs on Chart-MRAG Bench. In our evaluation, keypoint-based Correctness and Coverage metrics were introduced to rigorously assess accuracy and comprehensiveness.
The results reveal that unified multimodal embedding retrieval methods, which rely on a single vector store, perform poorly in high-density chart scenarios. 
Furthermore, even with ground-truth retrieval, the best-performing Claude-3.5 Sonnet \cite{team2024gemini} only achieved 58.19 Correctness and 73.87 Coverage metrics, highlighting persistent challenges in text-chart multimodal reasoning. In summary, the contributions of this paper are: 

1) We present Chart-based MRAG, the first extension of MRAG to chart scenarios that introduces a new dimension for evaluating crossmodal reasoning in information-dense visual contexts. 

2) We propose CHARGE, an automated framework for generating QA pairs in real-world scenarios through a structured pipeline of keypoint extraction, verification, and generation.


3) We establish Chart-MRAG Bench based on CHARGE. It is a human-verified benchmark for chart-based MRAG, covering 8 scenarios, 8 question types, and 4,738 QA pairs, with a subset designed for multi-hop reasoning. 

4) We introduce two robust evaluation metrics to assess MRAG quality. Extensive experiments highlight the limitations of existing retrieval and generation methods in chart-centric tasks.

\begin{table}[t]
  \centering
  \resizebox{\columnwidth}{!}{%
  \begin{tabular}{lcccc}
  \toprule
  \textbf{Benchmarks} & 
  \makecell{\textbf{Target}\\\textbf{Task}} &
  \makecell{\textbf{Retrieval}\\\textbf{Modality}} &
  \makecell{\textbf{Question}\\\textbf{Types}} & 
  \makecell{\textbf{Human}\\\textbf{Annotation}} \\
  \midrule
  OK-VQA ~\cite{marino2019ok} & VQA & Visual & 1 & $\checkmark$ \\
  MMQA ~\cite{talmor2021multimodalqa} & VQA & Multimodal & 16 & $\times$ \\
  PlotQA ~\cite{Methani_2020_WACV} & VQA & Visual & 1 & $\checkmark$ \\
  ChartQA ~\cite{masry2022chartqa} & VQA & Visual & 1 & $\checkmark$ \\
  DocVQA ~\cite{mathew2021docvqa} & VQA & Visual & 9 & $\checkmark$ \\
  MRAG-Bench ~\cite{hu2024mragbench} & MRAG & Visual & 3 & $\checkmark$ \\
  SSMQG ~\cite{wu2024synthetic} & MRAG & Multimodal & 5 & $\times$ \\
  \midrule
  \textbf{Chart-MRAG} \textit{(Ours)} & MRAG & Multimodal & 8 & $\checkmark$ \\ 
  \bottomrule
  \end{tabular}%
  }
  \caption{Comparison between existing MRAG benchmarks and the proposed Chart-MRAG Bench.}
  \label{tab:data_compare}
\end{table}

\section{Related Work}
\noindent \textbf{Multimodal RAG Methods.} Recent advances in Retrieval-Augmented Generation (RAG) \cite{izacard2022few,zhang2024raft} have successfully extended to multimodal domains \cite{chen2022murag,zhao2023retrieving,zhao2024retrieval}, enabling crossmodal tasks through MLLMs \cite{yao2024minicpm, sailvl}. While researchers have proposed various approaches \cite{ma-etal-2024-unifying,faysse2024colpali,yu2024visrag,Methani_2020_WACV,mathew2021docvqa} for crossmodal retrieval, current evaluation methodologies predominantly rely on Visual Question Answering (VQA) datasets \cite{marino2019ok,talmor2021multimodalqa,schwenk2022okvqa,masry2022chartqa}. These evaluations fall short in addressing retrieval-specific challenges.

\begin{figure*}[th]
  \includegraphics[width=\textwidth]{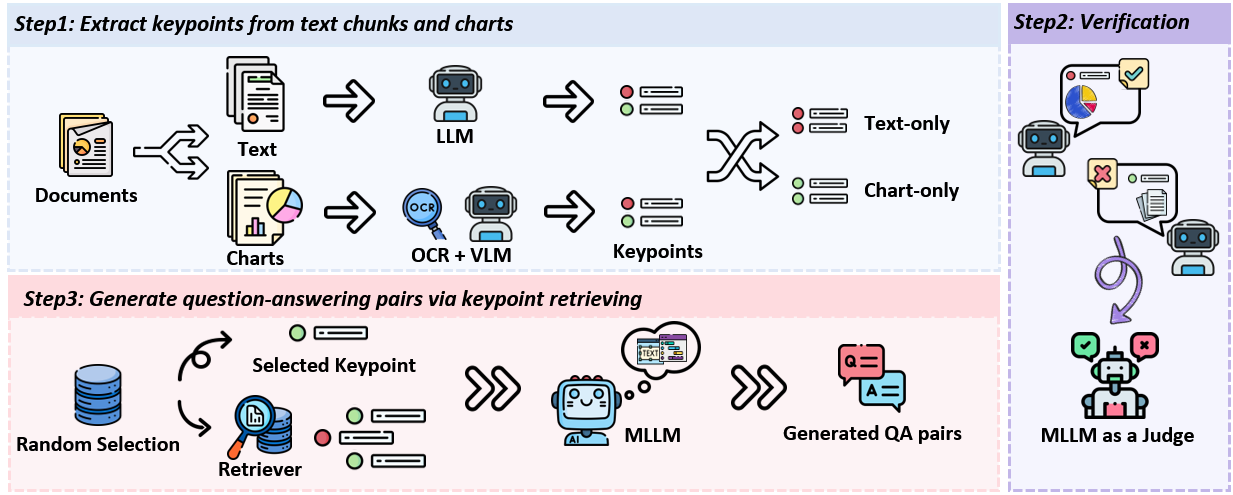}
  \caption{The proposed CHARGE framework for creating multimodal QA pairs from document-chart data, consisting of three steps: (1) Extract keypoints from textual content and charts, (2) Perform crossmodal verification to validate keypoint modality uniqueness, (3) Generate diverse QA pairs through constrained keypoint retrieval.}
  \label{Main_flow_diagram}
\end{figure*}

\noindent \textbf{Multimodal RAG Benchmarks.} The effectiveness of MRAG systems necessitates comprehensive evaluation benchmarks. While several benchmarks \cite{hu2024mragbench,li2024benchmarking,zhou2024megapairs} explore vision-based retrieval for question answering through manual annotation, they neglect the critical dimension of crossmodal collaborative generation. Some studies \cite{dong2025mmdocir,ma2024mmlongbench,ding2024mvqa} consider hybrid modality retrieval, yet they primarily rely on manual question-answering. Furthermore, although some studies \cite{es2023ragas,abaskohi2024fm2ds,mathew2021docvqa,li2024multimodal,wu2024synthetic} have investigated automated processes for generating crossmodal QA pairs, their scope focus on simplistic natural images with singular subjects, the chart-based scenarios largely unexplored. To bridge this gap, this paper introduce Chart-MRAG Bench. Table~\ref{tab:data_compare} illustrates the differences between existing MRAG benchmarks and Chart-MRAG Bench.
\section{CHARGE Framework}

We present CHARGE, a framework for generating multimodal multi-hop QA pairs from text-chart documents. CHARGE operates in three stages: (1) extracting self-contained keypoints from both textual and visual content, (2) verifying the modality authenticity of extracted keypoints through crossmodal verification, and (3) generating diverse QA pairs by combining related keypoints across documents and modalities.

\subsection{Extract Keypoints}

As illustrated in Fig ~\ref{Main_flow_diagram}, given multimodal documents $D = \{d_1,...,d_n\}$, CHARGE process its textual content into coherent chunks $T = \{t_1,...,t_m\}$ and charts as discrete units $C = \{c_1,...,c_k\}$. We define keypoints as self-contained factual statements that capture core information from these source materials. These atomic units are extracted from both textual and visual content (e.g., "33\% of U.S. adults say they use TikTok") through:

\begin{equation}
  \label{eq1}
  K = \begin{cases}
    \phi_t(T) & \textit{for text} \\
    \phi_c(C, \psi(C)) & \textit{for chart},
  \end{cases}
\end{equation}
where $K = \{k_1,...,k_r\}$ consists of structured information units capturing factual statements, logical inferences, or conclusive summaries. For textual content, we utilize \textrm{GPT-4o} through function $\phi_t$. For visual content, we first extract numerical values using function $\psi$ (implemented with \textit{ChartOCR} \cite{luo2021chartocr}), then employ \textrm{GPT-4o} through function $\phi_c$ to jointly process the charts $C$ and the extracted values, ensuring both contextual comprehension and numerical precision. Detailed workflow is presented in Appendix ~\ref{sec: keypoints_extraction}.

\subsection{Crossmodal Verification}
To ensure the reliability of extracted keypoints, we develop a crossmodal verification  mechanism that validates whether information truly belongs to its claimed modality. Our key insight is: Authentic modality-specific keypoints should be retrievable from its source modality but not from the other.

We first categorize keypoints into two fundamental types: (1) \textit{Text-based keypoints} ($K^T$): information exclusively present in textual form; (2) \textit{Chart-only keypoints} ($K^C$): information uniquely extractable from chart visualization. While \textit{GPT-4o} performs initial classification, crossmodal Verification is crucial for complex reasoning tasks.

The verification process employs crossmodal querying with \textit{GPT-4o} serving as a judge to determine whether the queried information exists in each modality's response. Taking text-based keypoint verification as an example, for a given keypoint $k^t_i \in K^T$, we query both its source text chunk $t_i$ and the paired chart $c_i$ (with OCR information $v_i$). Let $\tilde{k}^t_i$ and $\tilde{k}^c_i$ denote the model's responses from text and chart modalities respectively. The verification criterion is formalized as:

\begin{equation}
  \label{eq:verification}
  \text{Status}(k^t_i)\!=\!\begin{cases}
    \text{Retain} & \!\!\textit{if }\tilde{k}^t_i\!=\!k^t_i \land \tilde{k}^c_i\!\neq\!k^t_i \\
    \text{Drop} & \!\!\textit{otherwise}.
  \end{cases}
\end{equation}

This automated process retains keypoints only when correctly retrieved from their source modality and absent in others. Detailed algorithms are provided in Appendix ~\ref{sec:Crossmodal_algorithms}.

\begin{figure}[tbp]
  \includegraphics[width=\columnwidth]{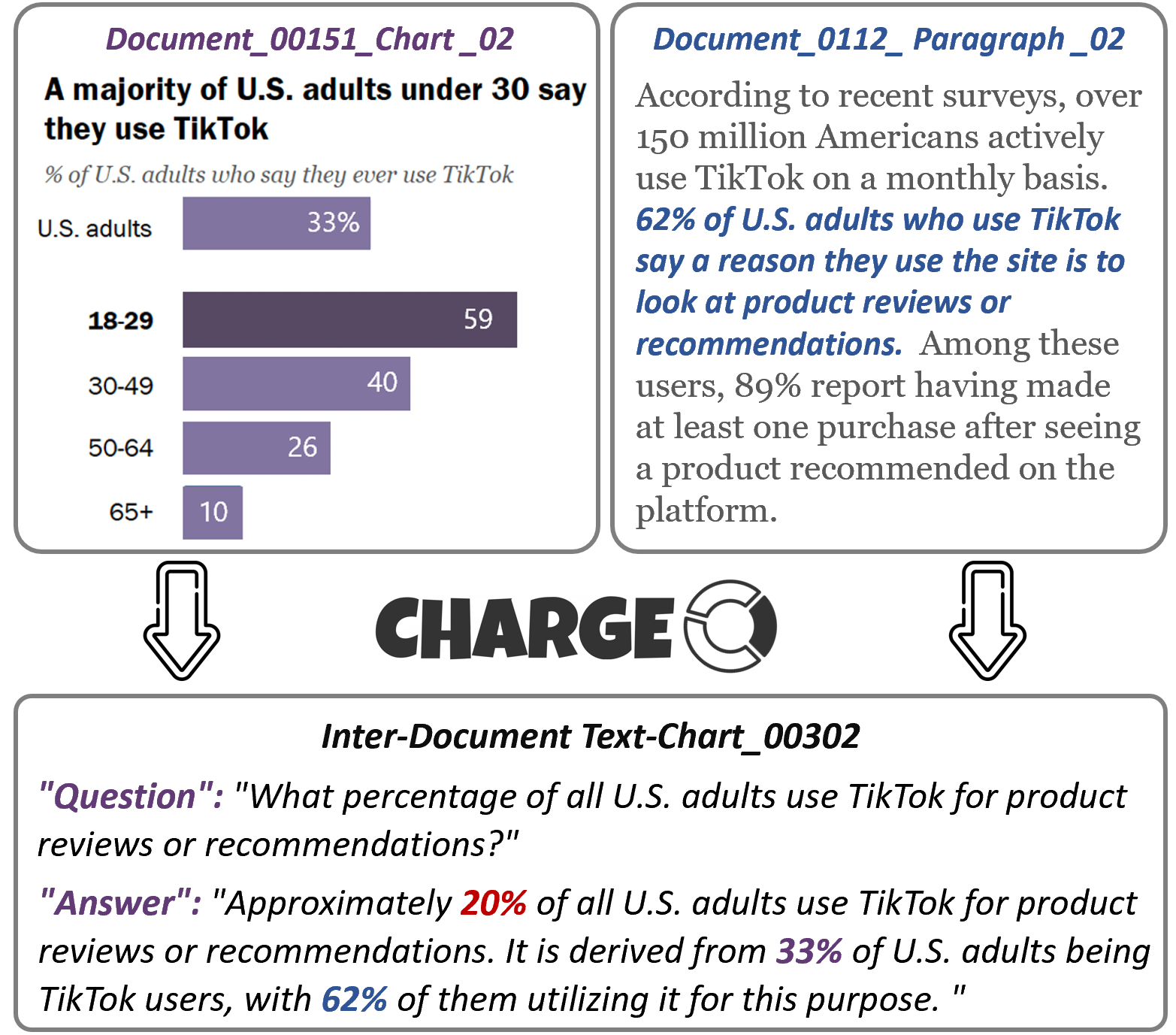}
  \caption{
    An inter-document multi-hop QA example from Chart-MRAG Bench, generated by CHARGE.
  }
  \label{case_diagram}
\end{figure}

\subsection{Question-Answer Pair Generation}

Since each keypoint represents a specific conclusion or data point, it can generate a corresponding question-answer pair (commonly referred to as Single-Point QA). In our CHARGE framework, we support this basic form of QA generation. Additionally, recognizing that most real-world queries require the integration of multiple knowledge points to be answered fully (known as Multi-hop QA), we further designed a multi-hop question answering approach: by combining semantically related keypoints to form a single question-answer pair. These types of questions cannot be completely answered using just one keypoint; they require the retrieval of all information sources (text chunks or charts) containing the constituent keypoints to be answered correctly. As illustrated in Fig~\ref{case_diagram}, CHARGE generates Multi-hop QA that requires retrieving multiple pieces of information by combining "33\% of U.S. adults say they use TikTok" (from a chart) with "62\% of U.S. adults who use TikTok say a reason they use the site is to look at product reviews or recommendations" (from text).

The generation of Multi-hop QA involves two key steps: identifying semantically related keypoints and constructing QA pairs from their combinations. Motivated by the capacity of RAG, we generate QA pairs through keypoint retrieval . Specifically, we first randomly select a keypoint (termed \textit{selected keypoint}) as the query, then retrieve candidate keypoints (termed \textit{retrieved keypoints}) by \textit{E5-Large} \cite{wang2022text}. By tracking the source documents of these retrieved keypoints, we categorize them into two types:
\begin{itemize}[noitemsep,leftmargin=*]
\item \textit{Intra-Document}: retrieved keypoints originate from the same document as the selected keypoint.
\item \textit{Inter-Document}: retrieved keypoints come from different documents than the selected keypoint.
\end{itemize}

Additionally, by tracking whether keypoints are from text or chart, we categorize their combinations into three types: 
\begin{itemize}[noitemsep,leftmargin=*]
\item \textit{Text-only}: both extracted from text paragraphs.
\item \textit{Chart-only}: both extracted from charts.
\item \textit{Text-Chart}: extracted from different modalities.
\end{itemize}

\begin{algorithm}[bp]
\SetAlgoLined
\DontPrintSemicolon
\small
\SetKwInOut{Input}{Input}
\SetKwInOut{Output}{Output}
\Input{Document set $D$; \\
       Chart keypoint set $K^C$; \\
       Text keypoint set $K^T$}
\Output{Question-Answer pair $(q, a)$}

\tcp{Step 1: Select chart keypoint}
Select $k^c_i \in K^C$ from document $d_a \in D$ \;
$(c_i, v_i) \gets (c, \psi(c))$ where $c \in d_a$ \;

\tcp{Step 2: Retrieve relevant text keypoint}
$K^T_r \gets \text{Retrieve}(k^c_i, K^T, k)$ \tcp*{top-k retrieval}
Select $k^t_j \in K^T_r$ from document $d_b \in D$ \;
$t_j \gets$ corresponding text block in $d_b$ \;

\tcp{Step 3: Generate QA pair}
$(q, a) \gets \text{MLLM}(k^c_i, k^t_j, c_i, v_i, t_j)$ \;

\Return{$(q, a)$}
\caption{\small Cross-document Text-Chart QA}
\label{alg:cross_doc_qa}
\end{algorithm}

For each combination of keypoints, we employ \textit{GPT-4o} to generate meaningful question-answer pairs by combining selected keypoint and its retrieved keypoint. These questions are designed to require multi-hop reasoning across different sources or modalities for answering. As illustrated in Table~\ref{tab:qa_types}, CHARGE naturally supports eight distinct types of QA pairs based on the document sources and modality combinations. While Algorithm~\ref{alg:cross_doc_qa} demonstrates the generation process for one specific type, comprehensive implementation details for all types are provided in Appendix~\ref{sec: qa_gen_algorithms}.

\begin{table}[t]
    \small
  \centering
  \setlength{\tabcolsep}{0pt}
  \begin{tabular*}{\columnwidth}
  {@{\extracolsep{\fill}}l@{\hspace{-0.5pt}}c@{\hspace{-0.5pt}}r@{}}
\toprule
{\small \textbf{Statistics}} & {\small \textbf{Reasoning Step}} & {\small \textbf{Number}} \\
\midrule
{\small Total questions} & {\small --} & {\small 4,738} \\
{\small ~- Single-Point Text-only} & {\small 1-hop} & {\small 499 (10.53\%)} \\
{\small ~- Single-Point Chart-only} & {\small 1-hop} & {\small 763 (16.10\%)} \\
{\small ~- Intra-Document  Text-only} & {\small 2-hop} & {\small 666 (14.06\%)} \\
{\small ~- Intra-Document Chart-only} & {\small 2-hop} & {\small 587 (12.39\%)} \\
{\small ~- Intra-Document Text-Chart} & {\small 2-hop} & {\small 746 (15.74\%)} \\
{\small ~- Inter-Document Text-only} & {\small 2-hop} & {\small 547 (11.54\%)} \\
{\small ~- Inter-Document Chart-only} & {\small 2-hop} & {\small 472 (9.96\%)} \\
{\small ~- Inter-Document Text-Chart} & {\small 2-hop} & {\small 458 (9.67\%)} \\
\bottomrule
\end{tabular*}
  \caption{Statistics of question types based on reasoning complexity and modality.}
  \label{tab:qa_types}
\end{table}

\section{Chart-MRAG Bench}

\begin{figure}[tbp]
    \centering
  \includegraphics[width=0.8\linewidth]{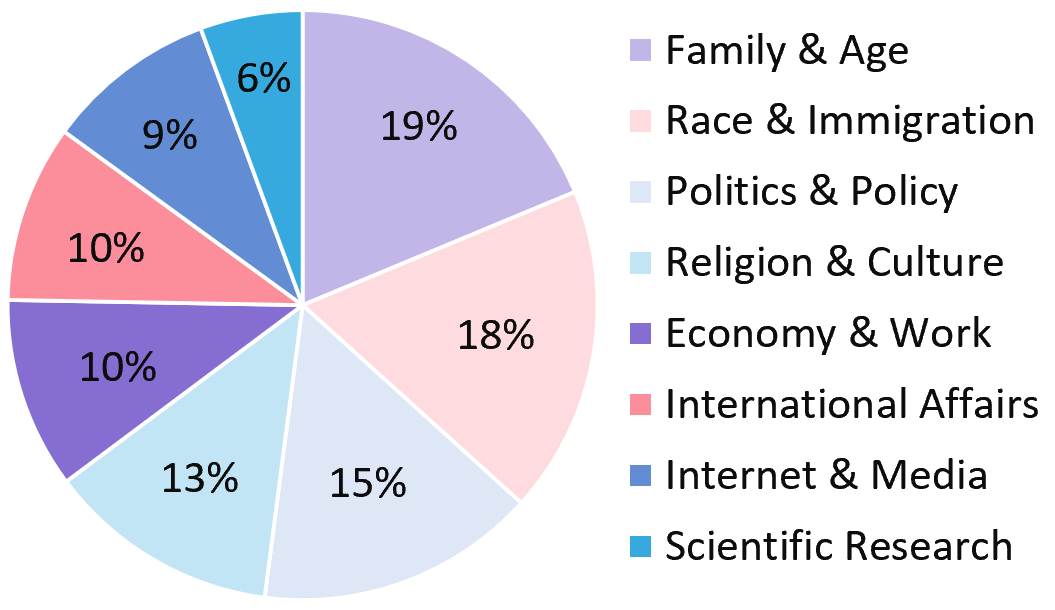}
  \caption{Distribution of documents across 8 domains, representing key areas of real-world applications.}
  \label{fig:theme_dist}
\end{figure}

By utilizing the CHARGE framework, we generated an initial pool of question-answer pairs. These pairs underwent rigorous expert evaluation to ensure high quality, culminating in the Chart-MRAG Bench. This process was guided by 4 principles:

\noindent \textbf{Authenticity and Diversity.} The benchmark is based on real-world data collected from the official website\footnote{www.pewresearch.org}, a trusted source of high-quality social research. We collected data from September 2023 to September 2024, encompassing 267 documents containing 1,283 text passages and 627 charts. As illustrated in Table~\ref{tab:qa_types} and Fig~\ref{fig:theme_dist}, Chart-MRAG Bench encompasses 8 distinct domains, integrating over 10 chart types and 8 QA types.

\noindent \textbf{Annotation Reliability.} We engaged 12 expert annotators with Master's degrees. All annotators were proficient in English, with an average TOEFL score of 92 or equivalent language proficiency. The annotation process took 34 working days to complete. Our annotation protocol involved three independent reviewers evaluating each sample, achieving a Fleiss's kappa ~\cite{fleiss1973equivalence} of 0.82, indicating substantial inter-annotator agreement.

\noindent \textbf{Rigorous Quality Control.} Through meticulous manual review, we refined the dataset from 9,600 initial candidates to 5,866 validated pairs by systematically eliminating 2,631 samples with OCR errors and 1,103 redundant samples. A consensus-based sampling strategy required validation from at least two reviewers, resulting in 4,738 high-quality samples (nearly 80\% of the validated pairs).

\noindent \textbf{High Information Complexity.} Statistical analysis reveals the benchmark's sophistication: approximately 70\% of charts contain more than 8 critical information points (mean: 13.87), and over 73\% of text passages include more than 6 keypoints (mean: 8.31). This information-rich environment rigorously evaluates models' capacity to process intricate and dense data representations.

For illustrative examples of Chart-MRAG Bench question-answer pairs across different domains and reasoning types, please refer to Appendix~\ref{sec: chart-mrag_bench_cases}.

\section{Experiments}

\input{retrieved_tables}
\input{generation_table}
\subsection{Baselines and Evaluation Metrics}
We conduct comprehensive evaluations using 3 distinct retrieval methods and 8 diverse MLLMs. Including 
\noindent \textbf{Multimodal Retrievers}: CLIP \cite{radford2021learning}, JINA \cite{koukounas2024jina}, SigLIP \cite{zhai2023sigmoid}, BGE-M3-base/large \cite{chen2024bge} and E5-base/large \cite{wang2022text}. 
\noindent And \textbf{Backbone MLLMs}: GPT-4o (version 2024-11-20) \cite{radford2021learning}, GPT-4-Vision \cite{radford2021learning}, Gemini-1.5-Pro \cite{team2024gemini}, Claude-3.5-Sonnet (version 2024-10-22) \cite{awadalla2023openflamingo}, SAIL-VL-2B \cite{sailvl}, Qwen2-VL-7B-instruct \cite{Qwen2VL}, MiniCPM-V-2.6 (8B) \cite{yao2024minicpm}, and Llama-3.2-90B-Vision \cite{dubey2024llama}.

Following \citet{wu2024synthetic}, we evaluate multimodal retrieval models using Recall@5 (R@5) and Recall@10 (R@10). Please refer to Appendix ~\ref{sec: retrieval_setup_and_metrics} for details of the retrieval setup and metrics.  Moreover, since chart-based MRAG is a newly proposed task, existing evaluation metrics are inadequate. Therefore, we introduce Correctness and Coverage metrics to assess the quality of responses.

\noindent \textbf{Correctness}. It measures the exact match between response and ground truth keypoints. Given a question-answer pair $\{Q, A, K^{gt}\}$ with ground truth keypoints $K^{gt} = \{k^{gt}_1, ..., k^{gt}_n\}$, we extract keypoints $K^r = \{k^r_1, ..., k^r_m\}$ from the model's response using an LLM. The score is defined as:
\begin{equation}
    \text{Correctness}(K^r, K^{gt}) = \mathbbm{1}[K^r \equiv K^{gt}],
\end{equation}
where $K^r \equiv K^{gt}$ implies complete keypoint matching and equal cardinality. This binary metric requires perfect accuracy, with zero tolerance for missing information or errors.

\noindent \textbf{Coverage}. It quantifies the proportion of correctly captured ground truth keypoints:
\begin{equation}
    \text{Coverage}(K^r, K^{gt}) = \frac{|K^{m}|}{|K^{gt}|},
\end{equation}
where $K^{m}$ represents matched ground truth keypoints. This continuous metric in [0,1] enables granular evaluation.

\subsection{Retrieval Performance Comparison}
Table~\ref{tab:retrieved_performance} reveals significant challenges in multimodal retrieval. While existing retrievers exhibit strong single-modal performance (JINA-CLIP achieves 77.78\% Recall@5 in text-only questions and SigLIP + E5 reaches 84.18\% Recall@5 in chart-only tasks), Inter-Document Text-Chart questions yielded only 22.32\% retrieval accuracy. The key findings demonstrate that storing and retrieving charts and text separately in the database substantially improves performance, achieving  recall rates of 42.53\% and 61.10\% at $k$=5 and $k$=10.

\noindent \textbf{Unified multimodal embeddings fail in knowledge-intensive scenarios.} While Method 1 outperforms all other approaches in pure text-only QA, it achieves zero recall (0.00\%) in chart-only QA and Text-Chart QA tasks. This phenomenon reveals a critical limitation: current unified multimodal embedding models excel at representing knowledge-sparse content (e.g., identifying a dog in an image) but struggle with knowledge-intensive scenarios (e.g., retrieving specific numerical values from charts in a multimodal repository).

\noindent \textbf{Chart captioning enables simple yet effective multimodal retrieval.} Methods 2 and 3 achieve comparable performance (Recall@5: 41.53\% vs 42.53\%), with differences primarily in chart retrieval due to the inherent limitations of text-based chart representations. However, considering the maintenance overhead of separate modal stores, caption-based retrieval provides a practical approach that preserves effectiveness while significantly reducing system complexity.

\subsection{Generative Performance Comparison}
Table~\ref{tab:generative_performance} presents the comprehensive experimental results of mainstream MLLMs, with retrieval method 3 consistently applied across all evaluations to ensure controlled comparison. The results reveal that state-of-the-art MLLMs achieve only modest performance metrics (Correctness = 3.06 and Coverage = 9.59) without multimodal RAG knowledge, highlighting Chart-MRAG Bench's exceptional challenging nature that surpasses existing benchmarks in knowledge leakage control. 

\noindent \textbf{Claude-3.5-Sonnet demonstrates superior overall performance.} The experimental results validate our keypoint-based evaluation methodology. With ground truth retrieval, Claude-3.5-Sonnet achieves Correctness of 58.19\% and Coverage of 73.87\%, outperforming mainstream MLLMs across various retrieval scenarios. It only falls behind Gemini-1.5-pro in Correctness at $k$=10. While Claude-3.5-Sonnet leads in aggregate scores, the narrow performance margins suggest potential in open-source alternatives: its Correctness (58.19) exceeds Gemini-1.5-pro by just 0.25. Moreover, in single-point text-only evaluation at recall $k$=5, qwen2-VL-7B-instruct achieves higher Correctness (50.51) compared to Claude-3.5-Sonnet (44.44).

\begin{figure}[th]
  \includegraphics[width=\columnwidth]{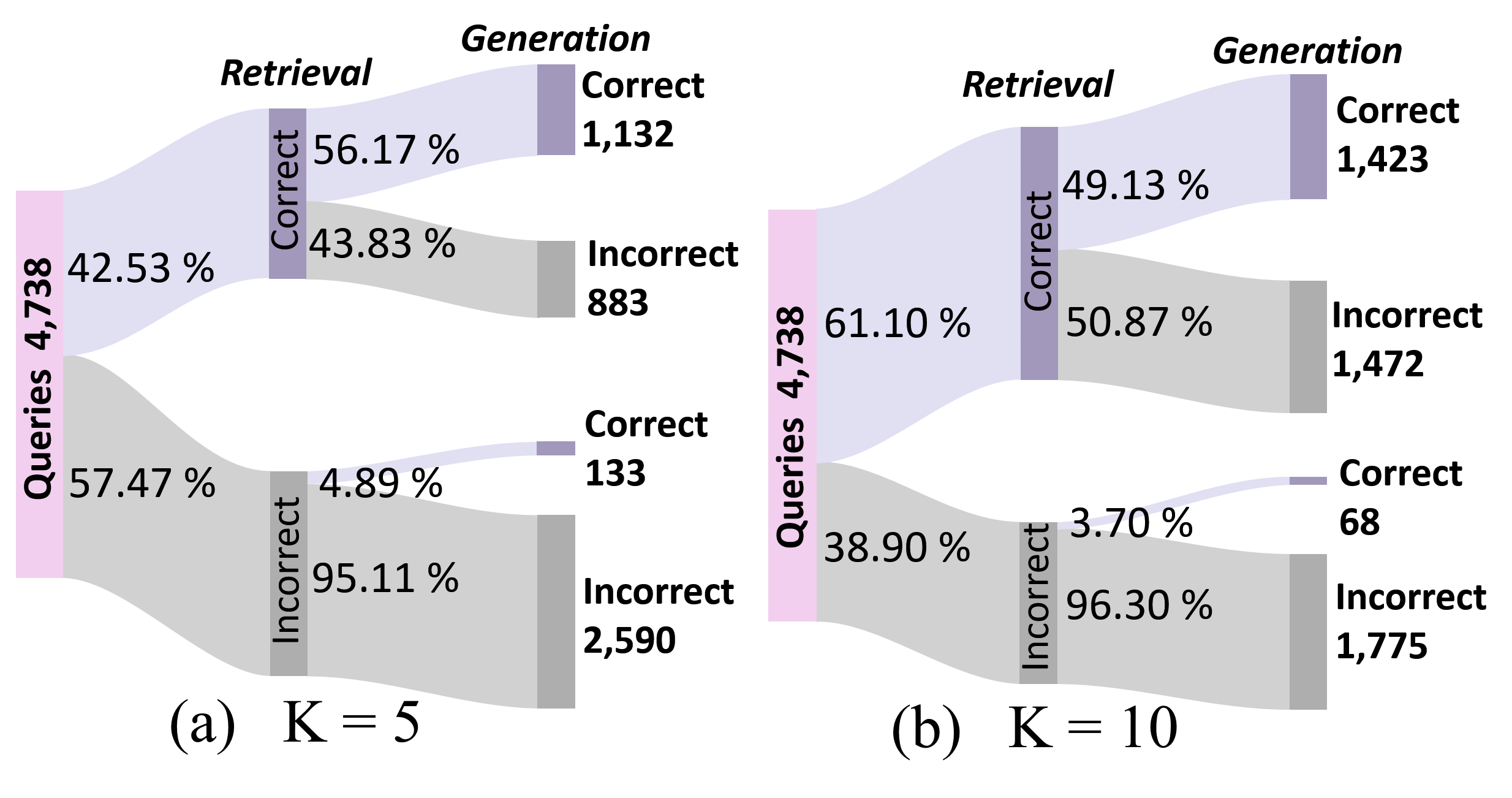}
  \caption{Impact of retrieval size $k$ across different parameter scales, demonstrating that larger models consistently benefit from increased retrieval context while smaller models show performance degradation.}
  \label{sankey_diagram}
\end{figure}

\noindent \textbf{Model performance generally scales with parameter count.} Among open-source MLLMs, Llama-3.2-90B-Vision consistently outperforms models with smaller parameters across various retrieval settings. Similarly, in proprietary MLLMs, GPT-4-Vision, with its presumably larger model size, demonstrates marginally better than GPT-4o.

\noindent \textbf{Architectural optimizations can mitigate MLLMs' parameter constraints.} By incorporating SigLip-400M and optimizing multi-image understanding, MiniCPM-V 2.6 achieves a Correctness of 46.94 and Coverage that surpasses its base model qwen2-VL-7B-instruct by 13.79 and 16.95 respectively. Most notably, despite using only 7B parameters, it approaches the performance of Llama-3.2-90B-Vision, with gaps of 3.22 in Correctness and 4.62 in Coverage, demonstrating that thoughtful architecture design can largely compensate for parameter constraints.
\begin{figure}[tbp]
  \includegraphics[width=\linewidth]{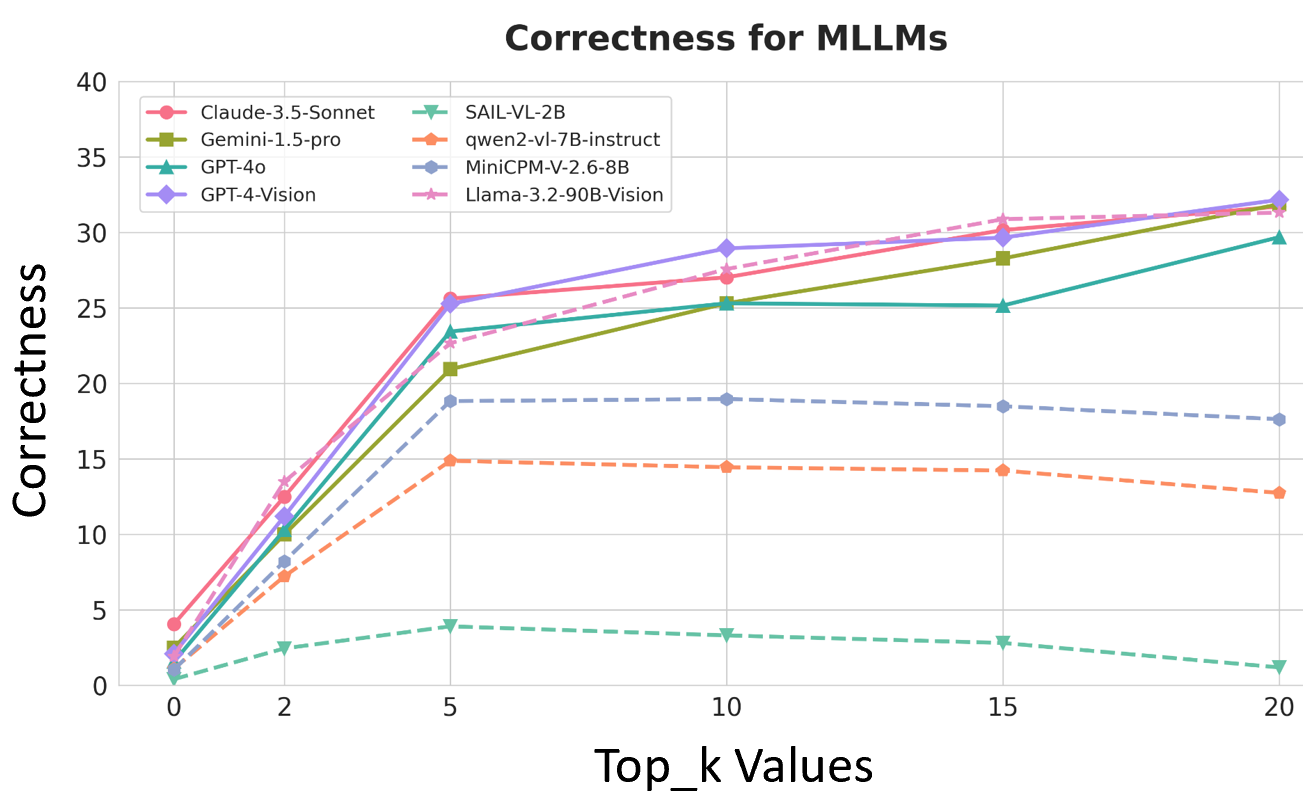}
  \caption{Trade-off analysis between retrieval coverage and answer accuracy across different $k$ settings, illustrating how larger retrieval windows increase recall while compromising answer correctness.}
  \label{retrieved_diagram}
\end{figure}
\subsection{Further Analysis}

In this study, we examine the influence of retrieval rate ($k$) and modality bias of MLLMs in multimodal question answering. Our analysis shows:

\noindent \textbf{Model performance in multimodal retrieval significantly correlates with parameter scale.} Empirical analysis reveals a strong correlation between model scale and multimodal retrieval performance. We evaluated eight models of varying parameter sizes under different retrieval settings ($k$ = 2, 5, 10, 15, 20), where retrieved items were balanced between images and text (split equally for even $k$, with text receiving one additional item for odd $k$). For each model, we selected 40 question-answer pairs per category, totaling 320 pairs for comprehensive evaluation, as shown in Fig \ref{retrieved_diagram}. The results demonstrate that larger models consistently achieve superior performance across all retrieval settings. In contrast, smaller models show no significant improvement (even exhibit declining) in performance as the number of retrieved items increases. 

\begin{figure}[t]
  \includegraphics[width=\linewidth]{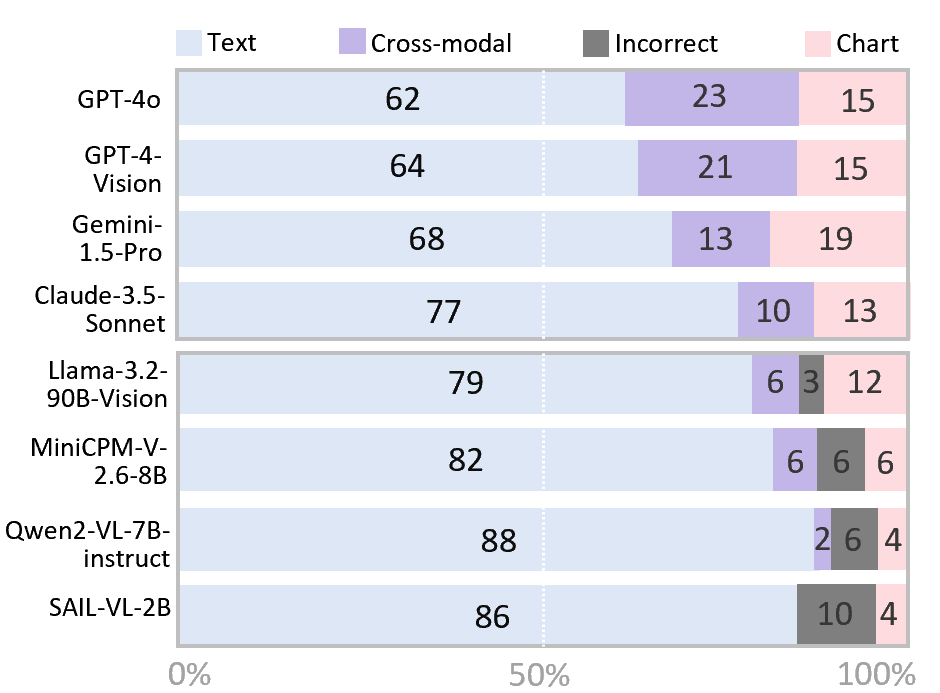}
  \caption{Analysis of modality preference in MLLMs when presented with redundant information across text and charts, revealing systematic modality bias.}
  \label{contrast_diagram}
\end{figure}

\noindent \textbf{Larger retrieval windows lead to a non-trivial trade-off between retrieval coverage and answer quality.} To systematically investigate the impact of Top\_$k$ on response generation, we conducted extended experiments as visualized in Fig ~\ref{sankey_diagram}. With $k$=5, the system achieves a 42.53 Recall and 56.17 correctness. When increasing $k$=10, although the 61.10 Recall, the answer get 49.13 correctness. Notably, while this adjustment results in an increase in absolute correct answers from 1,132 to 1,423, the improvement sacrifices precision.

\noindent \textbf{MLLMs demonstrate consistent text-over-visual modality bias.} To investigate modality bias in MLLMs, we carefully curated 100 specialized question-answer pairs where answers could be derived from both textual and visual information simultaneously, but with varying levels of granularity (e.g., "one third" in text versus "35.2\%" in charts). 
As shown in Fig.~\ref{contrast_diagram}, analysis reveals a consistent preference across models for text-only responses, even when charts contain more precise information. Notably, larger MLLMs demonstrate superior ability in detecting information redundancy and actively acknowledge this in their responses. For instance, GPT-4o proactively identified information redundancy in 23\% of its responses. In contrast, smaller models (such as SAIL-VL-2B) show limited sensitivity to such information redundancy. Detailed examples refer to Appendix ~\ref{sec: bias_case}.

\section{Conclusion}

 This paper introduces Chart-based MRAG, a novel task to bridge the evaluation gap for chart formats in MRAG systems. To support this, we propose CHARGE, an automated framework for generating Crossmodal evaluation samples with keypoint-based metrics. Combining CHARGE with expert validation, we construct Chart-MRAG Bench, comprising 4,738 high-quality question-answer pairs across 8 domains. Our experiments reveal key limitations in current MRAG approaches, highlighting the need for specialized architectures to better handle high-density visual interactions.


\section*{Limitations}
While our work presents promising results, we acknowledge several limitations that warrant consideration in future research.

First, although we ensured the accuracy of chart information in Chart-MRAG Bench through manual verification, the CHARGE framework would benefit from more advanced OCR techniques to further enhance the accuracy of question generation, especially in handling complex chart layouts and diverse visual elements.

Second, due to computational constraints, our evaluation was confined to a select set of MRAG methods and MLLMs. A more comprehensive evaluation across diverse model architectures and frameworks would likely yield additional insights into the generalizability of our findings and potentially reveal new directions for improvement.



\section*{Ethical Considerations}
This research was conducted under the approval of our institution's ethics review board. All procedures were designed to ensure participant welfare and data privacy throughout the study.

\paragraph{Participant Recruitment and Compensation.} We recruited expert annotators through Amazon, a professional data annotation platform. Annotators were compensated at a rate of \$28.5 per hour. This rate was determined by:
\begin{itemize}[noitemsep,leftmargin=*]
    \item Conducting pilot studies with 5 annotators to establish an average task completion time of 45 minutes
    \item Accounting for additional training time (30 minutes) and regular breaks
    \item Considering local living wage standards across different regions
    \item Adding a 20\% premium for specialized expertise required
\end{itemize}
For a typical 8-hour workday including training, we ensure fair payment while maintaining data quality. Regular feedback from annotators confirmed the compensation was considered fair for the required expertise and effort.

\paragraph{Informed Consent and Instructions.} All annotators received comprehensive instructions detailing the task requirements, data usage policies, and potential content exposure. The instruction package included:
\begin{itemize}[noitemsep,leftmargin=*]
    \item Task objectives and annotation guidelines
    \item Examples of expected annotations
    \item Data privacy and usage policies
    \item Right to withdraw from participation
\end{itemize}
Annotators provided explicit consent for their contributions to be used in academic research and public datasets.

\paragraph{Annotator Demographics.} Our annotation team consisted of 12 professional annotators with backgrounds in data science and visualization. The annotators represented diverse geographical locations (3 North America, 3 Europe, 6 Asia) and possessed relevant domain expertise. All demographic information was self-reported during the recruitment process.

\paragraph{Data Collection and Privacy.} The datasets used in this study, including those for generating multimodal question-answer pairs, were collected and processed in compliance with GDPR and relevant data privacy regulations. We ensured that:
\begin{itemize}[noitemsep,leftmargin=*]
    \item No personally identifiable information was collected
    \item All chart data was anonymized before annotation
    \item Participants were informed about data usage and sharing plans
\end{itemize}

\paragraph{Bias Mitigation.} We implemented several measures to minimize potential biases in our dataset and evaluation metrics:
\begin{itemize}[noitemsep,leftmargin=*]
    \item Diverse annotator selection to ensure varied perspectives
    \item Regular quality checks for systematic biases in annotations
    \item Balanced representation of different chart types and domains
\end{itemize}

The resulting benchmark will be made publicly available for academic research purposes, accompanied by detailed documentation of the collection process and annotator guidelines. All materials will be released through established academic repositories to ensure transparency and reproducibility.

\section*{Acknowledgements}
We sincerely thank all the anonymous reviewers. The work is supported by the National Natural Science Foundation of China (62206267 and 62176029), Chongqing Key Project of Technological Innovation and Application Development (CSTB2023TIAD-KPX0064), China Postdoctoral Science Foundation Funded Project (2024M763867).
\bibliography{custom}

\clearpage  
\appendix
\section{Keypoints Extraction Details} 
\label{sec: keypoints_extraction}

\begin{figure*}[hp]
  \includegraphics[width=\textwidth]{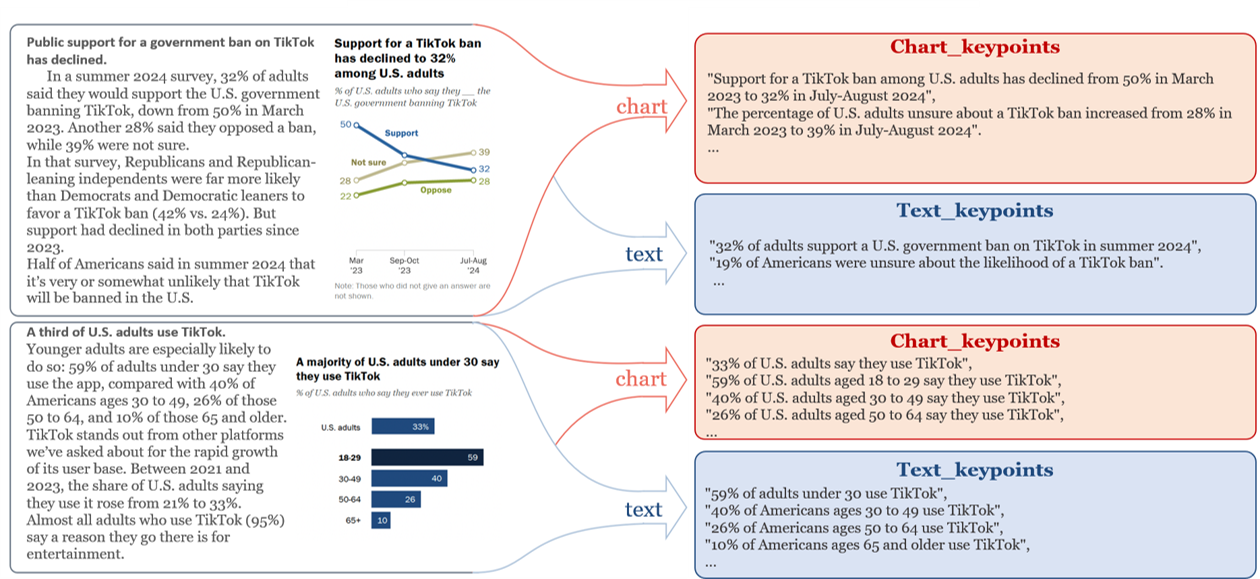}
  \caption{Demonstration of extracting atomic information units from text-chart pairs using structured prompts.}
  \label{keypoints_extraction_1}
\end{figure*}

As illustrated in Fig~\ref{keypoints_extraction_1}, we begin by extracting keypoints from text-chart pairs. Each keypoint represents an atomic unit that encapsulates a specific conclusive statement. The detailed extraction methodology is described in Appendix ~\ref{sec: prompts}.

\begin{figure*}[thbp]
  \includegraphics[width=\textwidth]{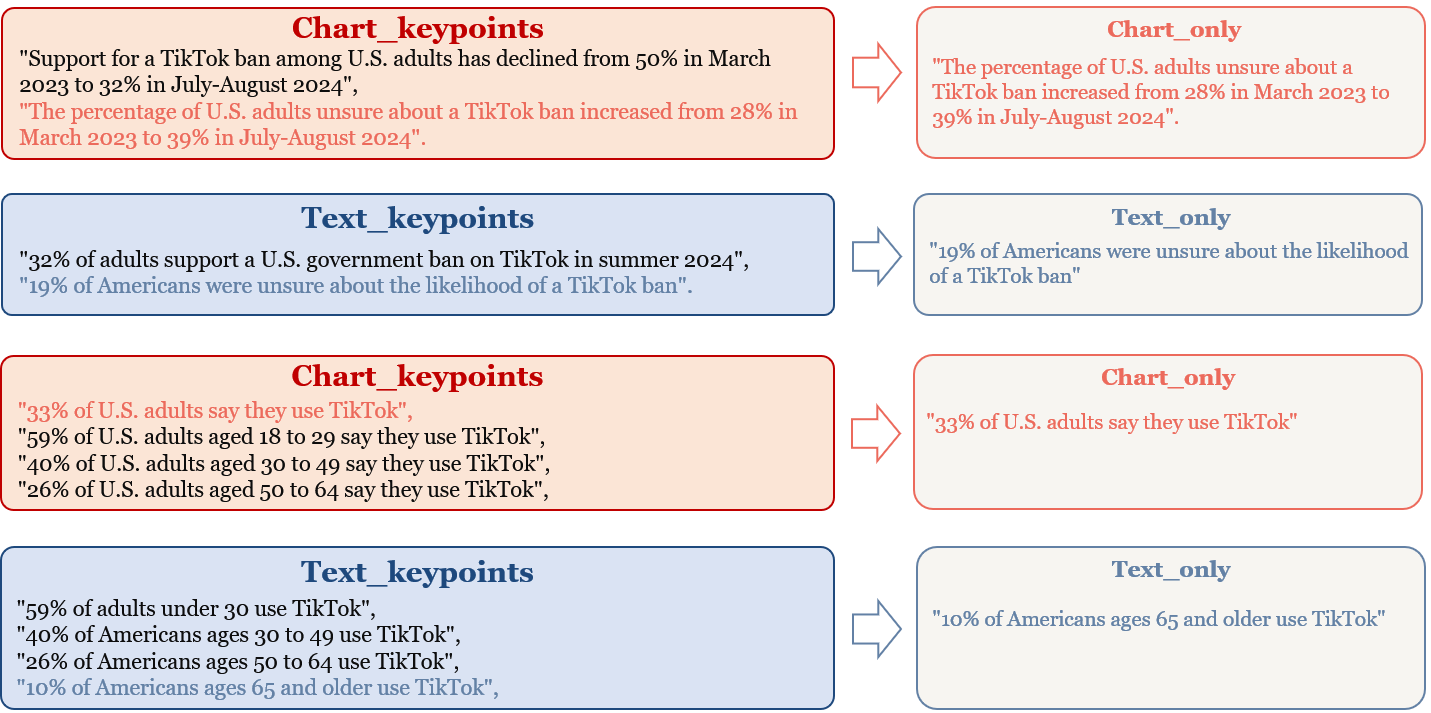}
  \caption{Illustration of the keypoints classification process using \textit{GPT-4o} screening and Crossmodal Verification.}
  \label{keypoints_extraction_2}
\end{figure*}

Following the initial extraction, as shown in Fig~\ref{keypoints_extraction_2}, we implement a two-stage filtering process (preliminary screening by \textit{GPT-4o} followed by Crossmodal Verification) to categorize the keypoints into two distinct sets:
\begin{itemize}[noitemsep,leftmargin=*]
\item \textit{Text-only keypoints} ($K^T$): information exclusively present in textual form
\item \textit{Chart-only keypoints} ($K^C$): information uniquely extractable from chart visualization
\end{itemize}

The detailed filtering methodology is provided in prompts~\ref{prompt1} and~\ref{prompt2}.

\section{Crossmodal Vertification Algorithms}
\label{sec:Crossmodal_algorithms}

Algorithm \ref{alg:text_only} presents a robust verification mechanism for text-only keypoint identification. The algorithm validates keypoints through Crossmodal Vertification, confirming a keypoint as text-only when text-only queries can yield the correct answer.
\begin{algorithm}[h]
\SetAlgoLined
\DontPrintSemicolon
\small
\SetKwInOut{Input}{Input}
\SetKwInOut{Output}{Output}
\Input{$K^T$: Text-only keypoints set, $k^t_i \in K^T$\\
       $T$: text chunks, $t_i \in T$\\
       $C$: chart set, $c_i \in C$\\
       $V$: chart info set, $v_i \in V$}
\Output{Updated $K^T$ with verification status}

\For{$k^t_i \in K^T$}{
    $q^t_i \gets LLM(k^t_i)$\;
    $\tilde{k}^t_i \gets LLM(q^t_i, t_i)$ \;
    $\tilde{k}^c_i \gets VLM(q^t_i, c_i, v_i)$ \;
    
    \tcp{Status determination}
    \uIf{$\tilde{k}^t_i = k^t_i$ \textbf{and} $\tilde{k}^c_i \neq k^t_i$}{
        $\text{Status}(k^t_i) \gets \text{Retain}$
    }
    \Else{
        $\text{Status}(k^t_i) \gets \text{Drop}$
    }
}
\Return{$K^T$}
\caption{\small Text-only Keypoint Verification}
\label{alg:text_only}
\end{algorithm}

Algorithm \ref{alg:chart_only} implements a symmetric verification approach for chart-only keypoint detection. Through inverse validation logic, it confirms keypoints as chart-only when only chart-only queries can produce the correct answer.

\begin{algorithm}[h]
\SetAlgoLined
\DontPrintSemicolon
\small
\SetKwInOut{Input}{Input}
\SetKwInOut{Output}{Output}
\Input{$K^C$: Chart-only keypoints set, $k^c_i \in K^C$\\
       $T$: text chunks, $t_i \in T$\\
       $C$: chart set, $c_i \in C$\\
       $V$: chart info set, $v_i \in V$}
\Output{Updated $K^C$ with verification status}

\For{$k^c_i \in K^C$}{
    $q^c_i \gets LLM(k^c_i)$ \;
    $\tilde{k}^t_i \gets LLM(q^c_i, t_i)$ \;
    $\tilde{k}^c_i \gets VLM(q^c_i, c_i, v_i)$ \;
    
    \tcp{Status determination}
    \uIf{$\tilde{k}^c_i = k^c_i$ \textbf{and} $\tilde{k}^t_i \neq k^c_i$}{
        $\text{Status}(k^c_i) \gets \text{Retain}$
    }
    \Else{
        $\text{Status}(k^c_i) \gets \text{Drop}$
    }
}
\Return{$K^C$}
\caption{\small Chart-only Keypoint Verification}
\label{alg:chart_only}
\end{algorithm}



\section{Question-Answering Pair Generation}
\label{sec: qa_gen_algorithms}
\paragraph{Single-Point Text-only QA} To generate single-point text-only question-answer pairs, we propose a simplified variant of the cross-document QA generation process. As shown in Algorithm~\ref{alg:single_point_text_qa}, the generation process consists of three main steps. First, we randomly select a text keypoint from the document collection that contains a complete, self-contained fact or statement. This approach focuses on validating discrete factual statements contained within a single text keypoint. The algorithm then leverages \textit{GPT-4o} to generate appropriate question-answer pairs based solely on the selected text keypoint and its context. This simplified approach ensures that the generated QA pairs require only single-hop reasoning, making them suitable for evaluating basic reading comprehension and fact extraction capabilities.

\begin{algorithm}[htbp]
\SetAlgoLined
\DontPrintSemicolon
\small
\SetKwInOut{Input}{Input}
\SetKwInOut{Output}{Output}
\Input{Document set $D$; \\
       Text keypoint set $K^T$}
\Output{Question-Answer pair $(q, a)$}

\tcp{Step 1: Select text keypoint}
Select $k^t_i \in K^T$ from document $d_a \in D$ \;
$t_i \gets$ corresponding text block in $d_a$ \;

\tcp{Step 2: Validate single-point constraint}
Assert $k^t_i$ contains complete fact or statement \;

\tcp{Step 3: Generate QA pair}
$(q, a) \gets \text{MLLM}(k^t_i, t_i)$ \;

\Return{$(q, a)$}
\caption{\small Single-Point Text-only QA}
\label{alg:single_point_text_qa}
\end{algorithm}

\paragraph{Single-Point Chart-only QA} For generating chart-focused question-answer pairs, we introduce a single-point variant that specializes in visual data comprehension. Algorithm~\ref{alg:single_chart_qa} begins by selecting a chart keypoint that represents a discrete observation from a visual element, such as a specific trend, comparison, or data point. Unlike Algorithm~\ref{alg:single_point_text_qa} which processes textual information, this approach extracts both the chart content and its corresponding numerical value to capture the complete visual context. The algorithm employs \textit{GPT-4o} to generate QA pairs that specifically test chart comprehension skills, ensuring that each question-answer pair is grounded in visual data interpretation without requiring cross-reference to textual content.

\begin{algorithm}[htbp]
\SetAlgoLined
\DontPrintSemicolon
\small
\SetKwInOut{Input}{Input}
\SetKwInOut{Output}{Output}
\Input{Document set $D$; \\
       Chart keypoint set $K^C$}
\Output{Question-Answer pair $(q, a)$}

\tcp{Step 1: Select chart keypoint}
Select $k^c_i \in K^C$ from document $d_a \in D$ \;
$c_i \gets$ chart content in $d_a$ \;
$v_i \gets$ corresponding value in $d_a$ \;

\tcp{Step 2: Validate single-point constraint}
Assert $k^c_i$ represents discrete chart observation \;

\tcp{Step 3: Generate QA pair}
$(q, a) \gets \text{MLLM}(k^c_i, c_i, v_i)$ \;

\Return{$(q, a)$}
\caption{\small Single-Point Chart-only QA}
\label{alg:single_chart_qa}
\end{algorithm}

\paragraph{Intra-Document Text-only QA} As illustrated in Algorithm~\ref{alg:intra_text_qa} introduces a systematic approach to constructing QA pairs that capture document-level semantic relationships. The algorithm first selects a primary text keypoint and its associated context, then identifies another relevant text keypoint from the same document to establish intra-document connections. This design ensures that the generated questions necessitate the integration of information from multiple parts of the document, testing a system's ability to perform document-level reasoning and information synthesis. The final generation step employs \textit{GPT-4o} to create questions that effectively evaluate comprehensive document understanding capabilities.
\begin{algorithm}[htbp]
\SetAlgoLined
\DontPrintSemicolon
\small
\SetKwInOut{Input}{Input}
\SetKwInOut{Output}{Output}
\Input{Document set $D$; \\
       Text keypoint set $K^T$}
\Output{Question-Answer pair $(q, a)$}

\tcp{Step 1: Select primary text keypoint}
Select $k^t_i \in K^T$ from document $d_a \in D$ \;
$t_i \gets$ corresponding text block in $d_a$ \;

\tcp{Step 2: Retrieve intra-document text keypoint}
$K^T_r \gets \{k^t \in K^T | k^t \text{ from } d_a\}$ \;
Select $k^t_j \in K^T_r$ \;
$t_j \gets$ corresponding text block in $d_a$ \;

\tcp{Step 3: Generate QA pair}
$(q, a) \gets \text{MLLM}(k^t_i, k^t_j, t_i, t_j)$ \;

\Return{$(q, a)$}
\caption{\small Intra-document Text-only QA}
\label{alg:intra_text_qa}
\end{algorithm}

\paragraph{Intra-Document Chart-only QA} Building upon our text-only approach, we propose an algorithm that focuses on reasoning across multiple chart elements within a single document. Algorithm~\ref{alg:intra_chart_qa} presents a structured methodology for generating questions that require the synthesis of information from related visual components. The algorithm initiates by selecting a primary chart keypoint and extracts its corresponding visual features, then identifies semantically related chart elements within the same document to establish comprehensive visual reasoning chains. This architecture enables the generation of questions that assess a system's capability to integrate and reason over multiple visual representations while maintaining document-level consistency. The generation process leverages \textit{GPT-4o} to construct questions that effectively evaluate sophisticated chart comprehension and cross-reference abilities.
\begin{algorithm}[htbp]
\SetAlgoLined
\DontPrintSemicolon
\small
\SetKwInOut{Input}{Input}
\SetKwInOut{Output}{Output}
\Input{Document set $D$; \\
       Chart keypoint set $K^C$}
\Output{Question-Answer pair $(q, a)$}

\tcp{Step 1: Select primary chart keypoint}
Select $k^c_i \in K^C$ from document $d_a \in D$ \;
$(c_i, v_i) \gets (c, \psi(c))$ where $c \in d_a$ \;

\tcp{Step 2: Retrieve intra-document chart keypoint}
$K^C_r \gets \{k^c \in K^C | k^c \text{ from } d_a\}$ \;
Select $k^c_j \in K^C_r$ \;
$(c_j, v_j) \gets (c, \psi(c))$ where $c \in d_a$ \;

\tcp{Step 3: Generate QA pair}
$(q, a) \gets \text{MLLM}(k^c_i, k^c_j, c_i, v_i, c_j, v_j)$ \;

\Return{$(q, a)$}
\caption{\small Intra-document Chart-only QA}
\label{alg:intra_chart_qa}
\end{algorithm}

\paragraph{Intra-Document Text-Chart QA} To further enhance the document-level reasoning capabilities, we introduce an algorithm that bridges the gap between textual and visual content within individual documents. Algorithm~\ref{alg:intra_text_chart_qa} establishes a novel approach by first selecting chart-specific keypoints and then retrieving semantically related textual descriptions from the same document. This design facilitates the generation of questions that require joint understanding of both modalities, particularly focusing on how charts and their contextual textual explanations complement each other. Through semantic retrieval between chart and text keypoints, the algorithm ensures that the generated questions capture authentic Crossmodal relationships while maintaining document coherence. The generation process employs \textit{GPT-4o} to synthesize questions that evaluate systems' ability to perform integrated reasoning over both visual and textual information sources.

\begin{algorithm}[htbp]
\SetAlgoLined
\DontPrintSemicolon
\small
\SetKwInOut{Input}{Input}
\SetKwInOut{Output}{Output}
\Input{Document set $D$; \\
       Chart keypoint set $K^C$; \\
       Text keypoint set $K^T$}
\Output{Question-Answer pair $(q, a)$}

\tcp{Step 1: Select chart keypoint}
Select $k^c_i \in K^C$ from document $d_a \in D$ \;
$(c_i, v_i) \gets (c, \psi(c))$ where $c \in d_a$ \;

\tcp{Step 2: Retrieve intra-document text keypoint}
$K^T_r \gets \{k^t \in K^T | k^t \text{ from } d_a\}$ \;
$k^t_j \gets {k^t \in K^T_r} \text{Similarity}(k^c_i, k^t)$ \;
$t_j \gets$ corresponding text block in $d_a$ \;

\tcp{Step 3: Generate QA pair}
$(q, a) \gets \text{MLLM}(k^c_i, k^t_j, c_i, v_i, t_j)$ \;

\Return{$(q, a)$}
\caption{\small Intra-document Text-Chart QA}
\label{alg:intra_text_chart_qa}
\end{algorithm}

\paragraph{Inter-Document Text-only QA} To extend our intra-document approach to a broader context, we develop an algorithm that enables reasoning across different documents. Algorithm~\ref{alg:inter_text_qa} introduces a systematic methodology for generating questions that require the integration of information from multiple source documents. The algorithm first selects a primary text keypoint from one document, then identifies semantically related text content from a different document, thereby establishing cross-document connections. This design facilitates the generation of questions that assess a system's ability to synthesize information across document boundaries while maintaining logical coherence. The generation process employs \textit{GPT-4o} to create questions that effectively evaluate comprehensive cross-document understanding and reasoning capabilities.

\begin{algorithm}[htbp]
\SetAlgoLined
\DontPrintSemicolon
\small
\SetKwInOut{Input}{Input}
\SetKwInOut{Output}{Output}
\Input{Document set $D$; \\
       Text keypoint set $K^T$}
\Output{Question-Answer pair $(q, a)$}

\tcp{Step 1: Select primary text keypoint}
Select $k^t_i \in K^T$ from document $d_a \in D$ \;
$t_i \gets$ corresponding text block in $d_a$ \;

\tcp{Step 2: Retrieve cross-document text keypoint}
$K^T_r \gets \{k^t \in K^T | k^t \text{ from } d_b \in D, d_b \neq d_a\}$ \;
Select $k^t_j \in K^T_r$ \;
$t_j \gets$ corresponding text block in $d_b$ \;

\tcp{Step 3: Generate QA pair}
$(q, a) \gets \text{MLLM}(k^t_i, k^t_j, t_i, t_j)$ \;

\Return{$(q, a)$}
\caption{\small Inter-document Text-only QA}
\label{alg:inter_text_qa}
\end{algorithm}

\paragraph{Inter-Document Chart-only QA} To further advance cross-document reasoning capabilities, we present an algorithm that enables sophisticated analysis across charts from different documents. Algorithm~\ref{alg:inter_chart_qa} establishes a methodology for generating questions that require the synthesis of visual information spanning multiple documents. The algorithm begins by selecting a primary chart keypoint and its visual features from one document, then identifies related chart elements from a different document to establish cross-document visual reasoning chains. This framework facilitates the generation of questions that evaluate a system's ability to integrate and reason over visual representations across document boundaries while maintaining semantic coherence. The generation process utilizes \textit{GPT-4o} to create questions that effectively assess advanced chart comprehension and cross-document visual reasoning abilities.

\begin{algorithm}[htbp]
\SetAlgoLined
\DontPrintSemicolon
\small
\SetKwInOut{Input}{Input}
\SetKwInOut{Output}{Output}
\Input{Document set $D$; \\
       Chart keypoint set $K^C$}
\Output{Question-Answer pair $(q, a)$}

\tcp{Step 1: Select primary chart keypoint}
Select $k^c_i \in K^C$ from document $d_a \in D$ \;
$(c_i, v_i) \gets (c, \psi(c))$ where $c \in d_a$ \;

\tcp{Step 2: Retrieve cross-document chart keypoint}
$K^C_r \gets \{k^c \in K^C | k^c \text{ from } d_b \in D, d_b \neq d_a\}$ \;
Select $k^c_j \in K^C_r$ \;
$(c_j, v_j) \gets (c, \psi(c))$ where $c \in d_b$ \;

\tcp{Step 3: Generate QA pair}
$(q, a) \gets \text{MLLM}(k^c_i, k^c_j, c_i, v_i, c_j, v_j)$ \;

\Return{$(q, a)$}
\caption{\small Inter-document Chart-only QA}
\label{alg:inter_chart_qa}
\end{algorithm}

\paragraph{Inter-Document Text-Chart QA} Extending our Crossmodal reasoning framework beyond single documents, we propose an algorithm that enables sophisticated analysis across textual and visual content from different documents. Algorithm~\ref{alg:inter_text_chart_qa} establishes an advanced approach by selecting chart-specific keypoints from one document and retrieving semantically related textual descriptions from another document. This architecture facilitates the generation of questions that require joint understanding of cross-document modalities, particularly exploring how charts and textual explanations from different sources can be synthesized for comprehensive reasoning. Through cross-document semantic retrieval between chart and text keypoints, the algorithm generates questions that evaluate systems' ability to perform integrated reasoning across both document boundaries and modality gaps. The generation process leverages \textit{GPT-4o} to create questions that assess sophisticated cross-document visual-textual understanding capabilities.

\begin{algorithm}[htbp]
\SetAlgoLined
\DontPrintSemicolon
\small
\SetKwInOut{Input}{Input}
\SetKwInOut{Output}{Output}
\Input{Document set $D$; \\
       Chart keypoint set $K^C$; \\
       Text keypoint set $K^T$}
\Output{Question-Answer pair $(q, a)$}

\tcp{Step 1: Select chart keypoint}
Select $k^c_i \in K^C$ from document $d_a \in D$ \;
$(c_i, v_i) \gets (c, \psi(c))$ where $c \in d_a$ \;

\tcp{Step 2: Retrieve cross-document text keypoint}
$K^T_r \gets \{k^t \in K^T | k^t \text{ from } d_b \in D, d_b \neq d_a\}$ \;
$k^t_j \gets {k^t \in K^T_r} \text{Similarity}(k^c_i, k^t)$ \;
$t_j \gets$ corresponding text block in $d_b$ \;

\tcp{Step 3: Generate QA pair}
$(q, a) \gets \text{MLLM}(k^c_i, k^t_j, c_i, v_i, t_j)$ \;

\Return{$(q, a)$}
\caption{\small Inter-document Text-Chart QA}
\label{alg:inter_text_chart_qa}
\end{algorithm}

\section{Chart-MRAG Bench Cases}
\label{sec: chart-mrag_bench_cases}
\begin{figure*}[htbp] 
  \includegraphics[width=\textwidth]{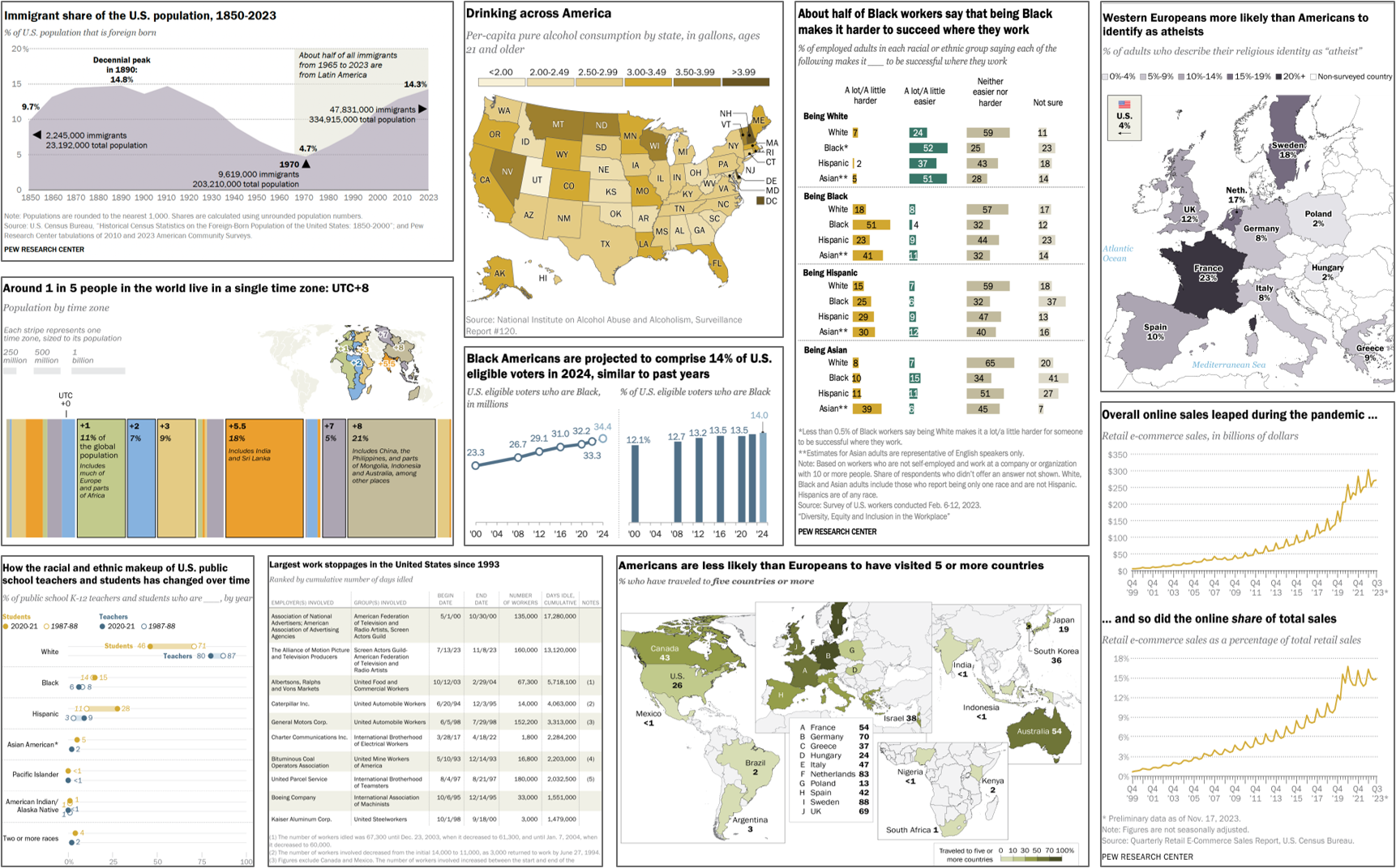}
  \caption{Representative visualization categories from Chart-MRAG Bench, showcasing temporal trend analysis (line charts), geospatial data visualization (choropleth maps), categorical comparisons (bar charts), compositional analysis (stacked bars), and integrated text-chart representations. The diversity of these examples demonstrates the comprehensive scope of Chart-MRAG Bench in representing complex statistical information across multiple domains and visualization paradigms.}
  \label{chart_categories}
\end{figure*}

To illustrate the diverse chart categories in Chart-MRAG Bench, we present representative examples as shown in Figure~\ref{chart_categories}.

\begin{figure*}[htbp] 
  \includegraphics[width=\textwidth]{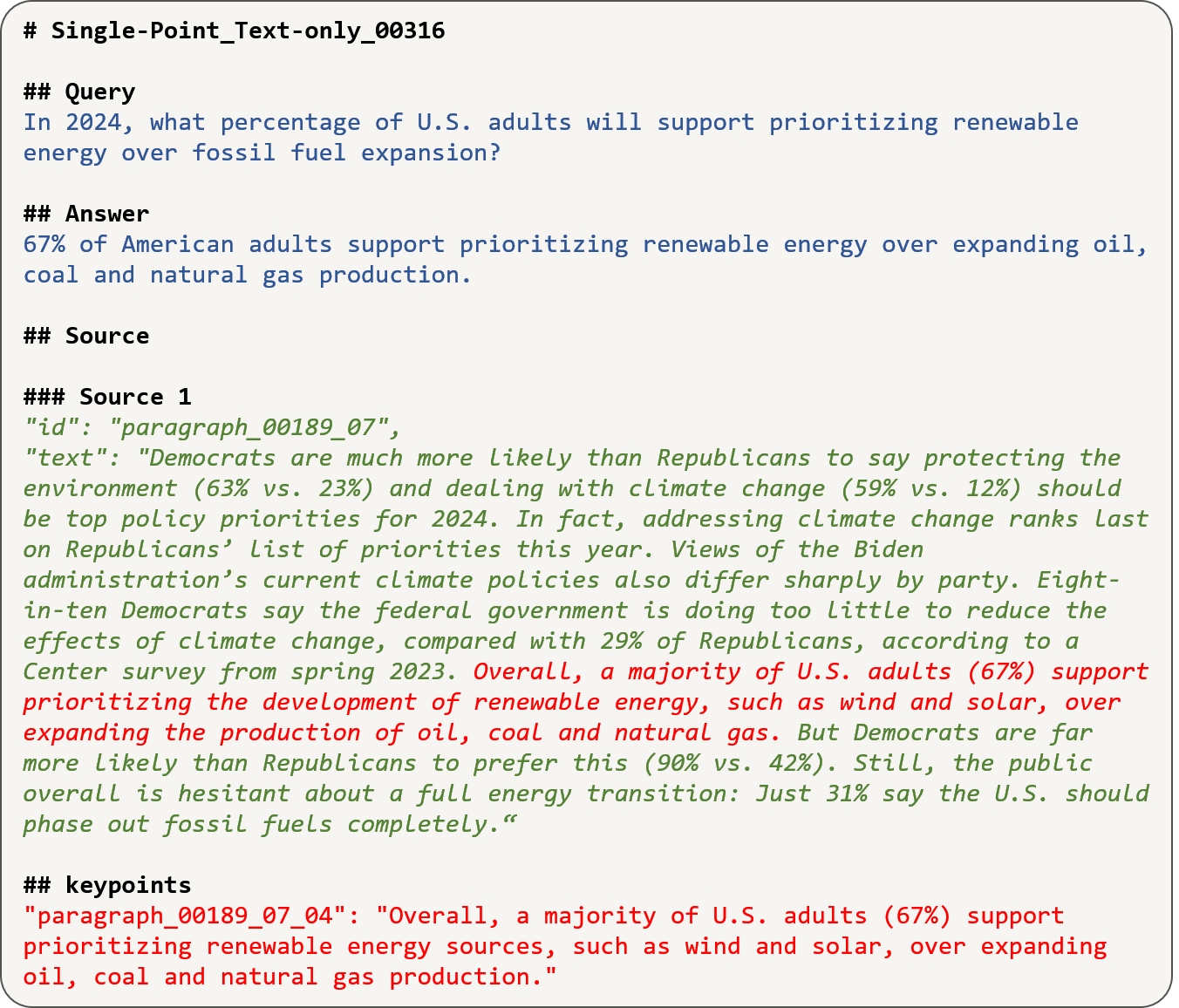}
  \caption{A sample case of single-point text-only question answering.}
  \label{sample_case_1}
\end{figure*}

\begin{figure*}[htbp] 
  \includegraphics[width=\textwidth]{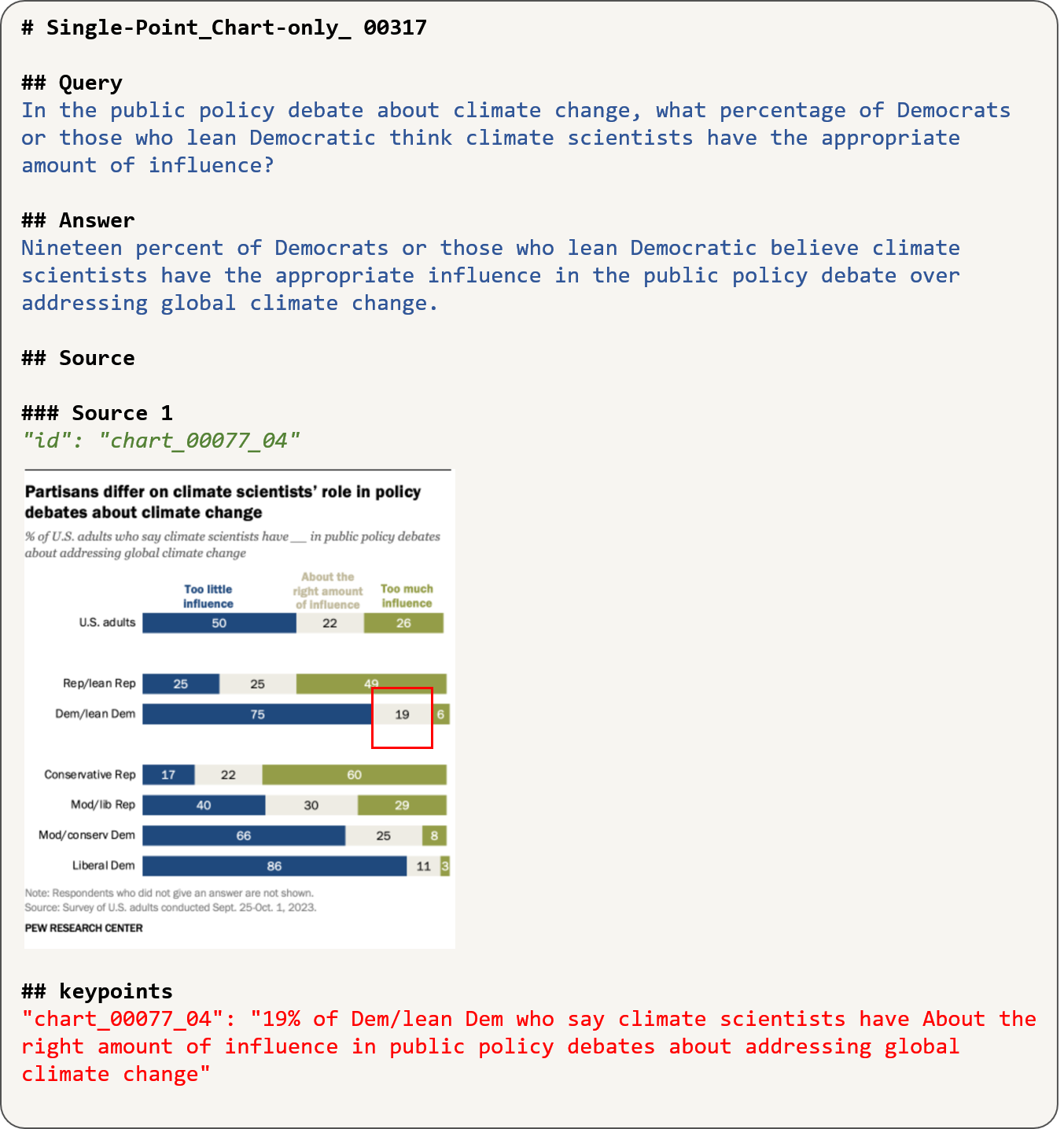}
  \caption{A sample case of single-point chart-only question answering.}
  \label{sample_case_2}
\end{figure*}

\begin{figure*}[htbp] 
  \includegraphics[width=\textwidth]{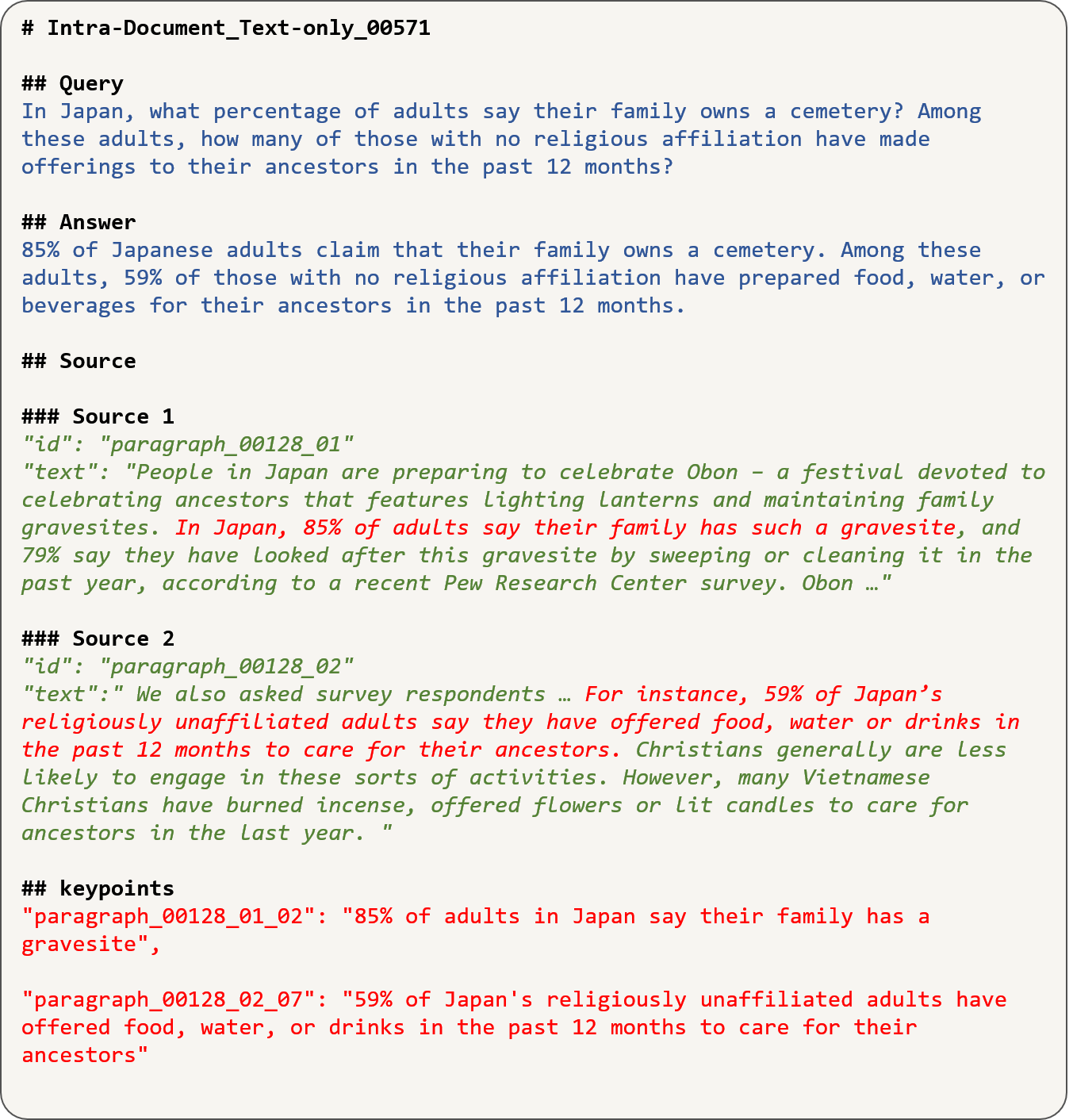}
  \caption{A sample case of intra-document text-only question answering.}
  \label{sample_case_3}
\end{figure*}

\begin{figure*}[htbp] 
  \includegraphics[width=\textwidth]{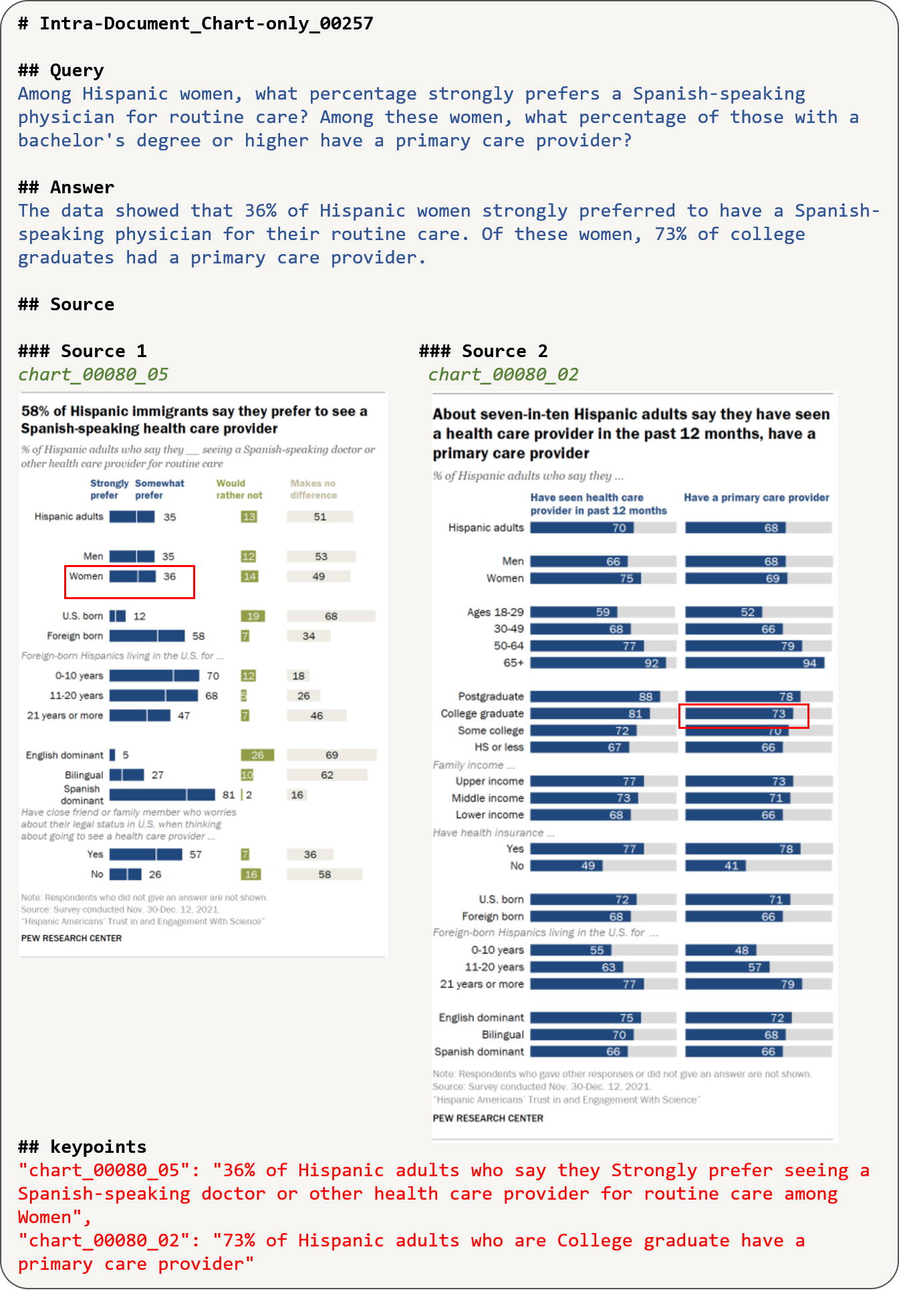}
  \caption{A sample case of intra-document chart-only question answering.}
  \label{sample_case_4}
\end{figure*}

\begin{figure*}[htbp] 
  \includegraphics[width=\textwidth]{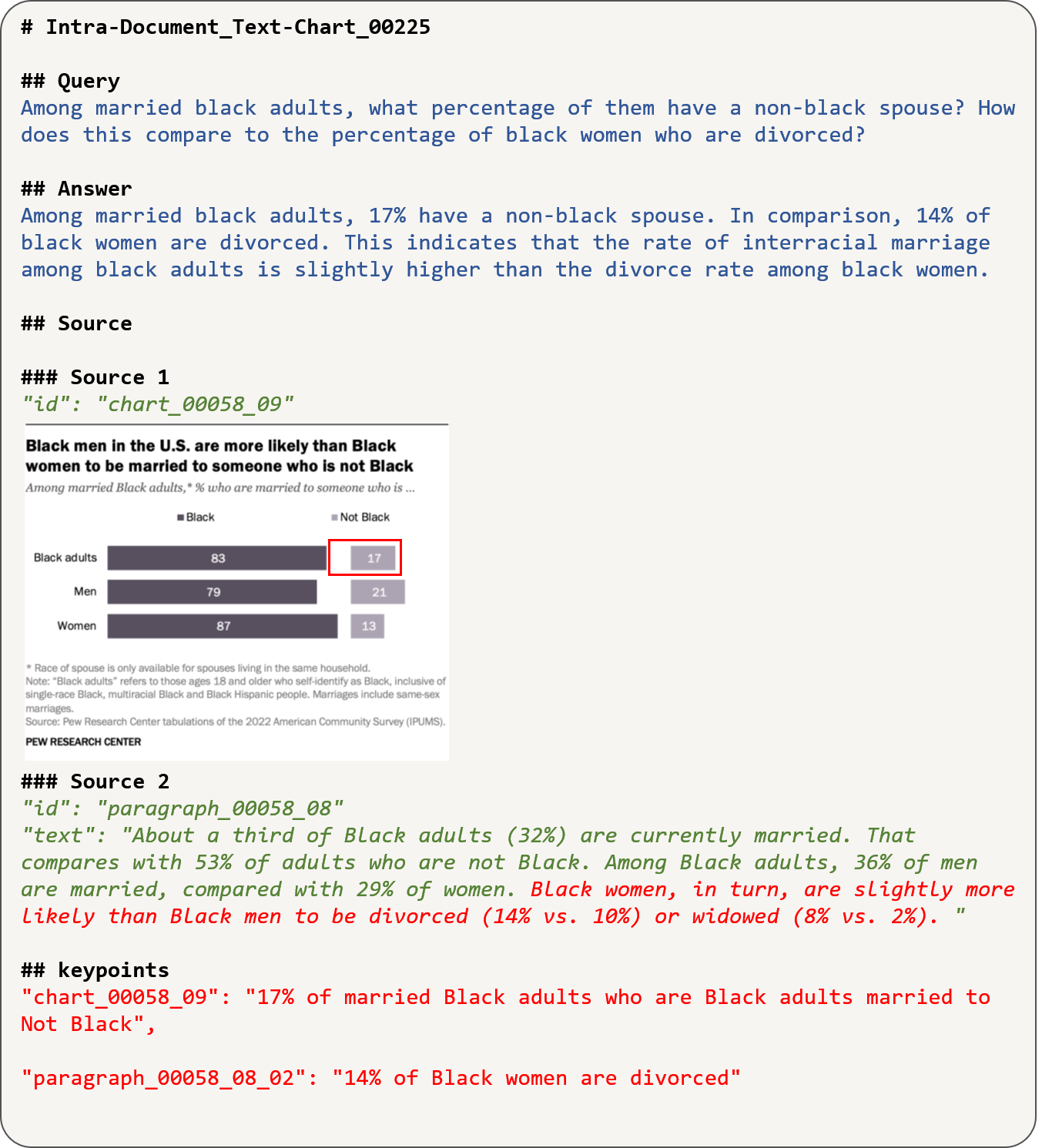}
  \caption{A sample case of intra-document text-chart question answering.}
  \label{sample_case_5}
\end{figure*}

\begin{figure*}[htbp] 
  \includegraphics[width=\textwidth]{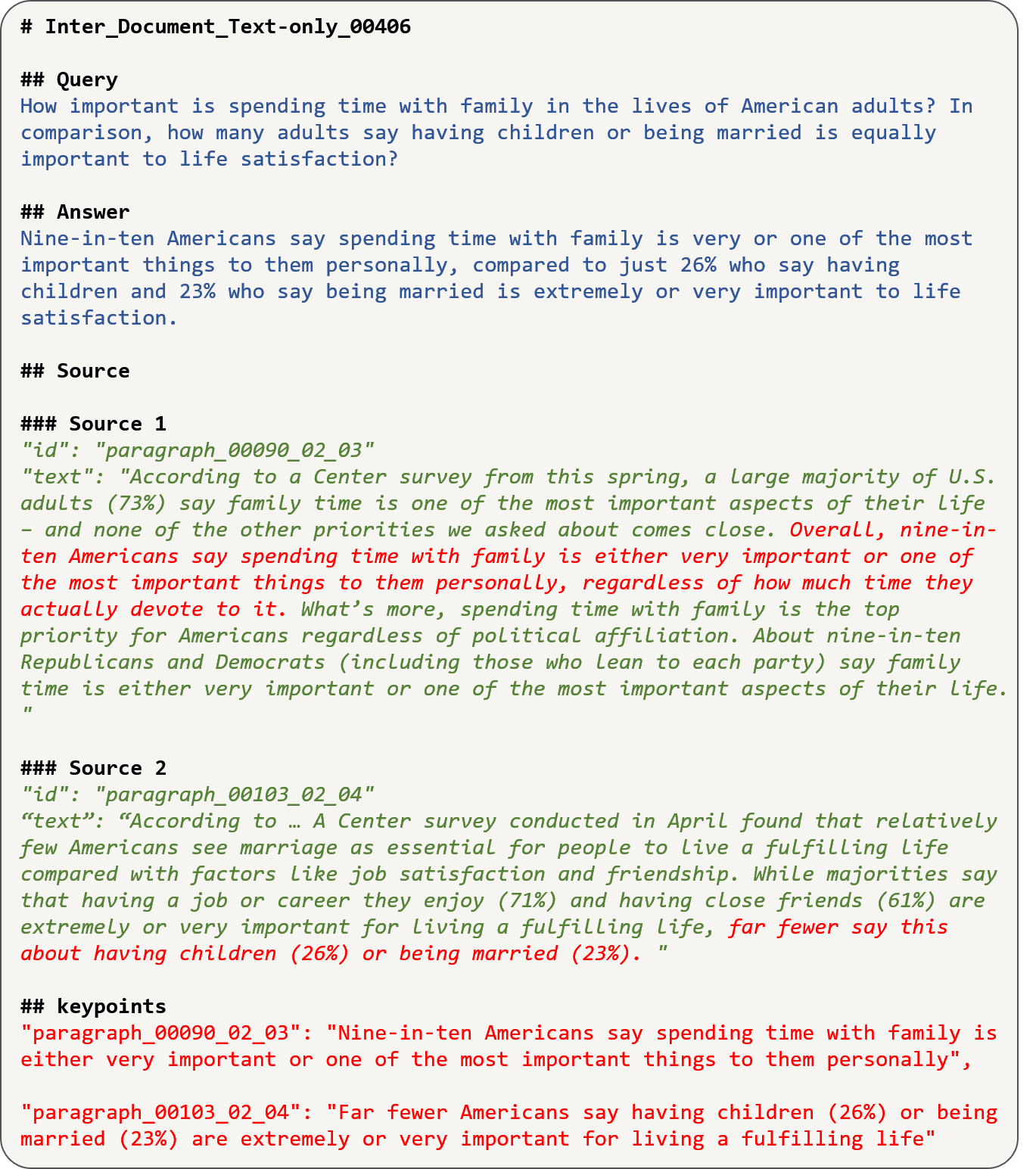}
  \caption{A sample case of inter-document text-only question answering.}
  \label{sample_case_6}
\end{figure*}

\begin{figure*}[htbp] 
  \includegraphics[width=\textwidth]{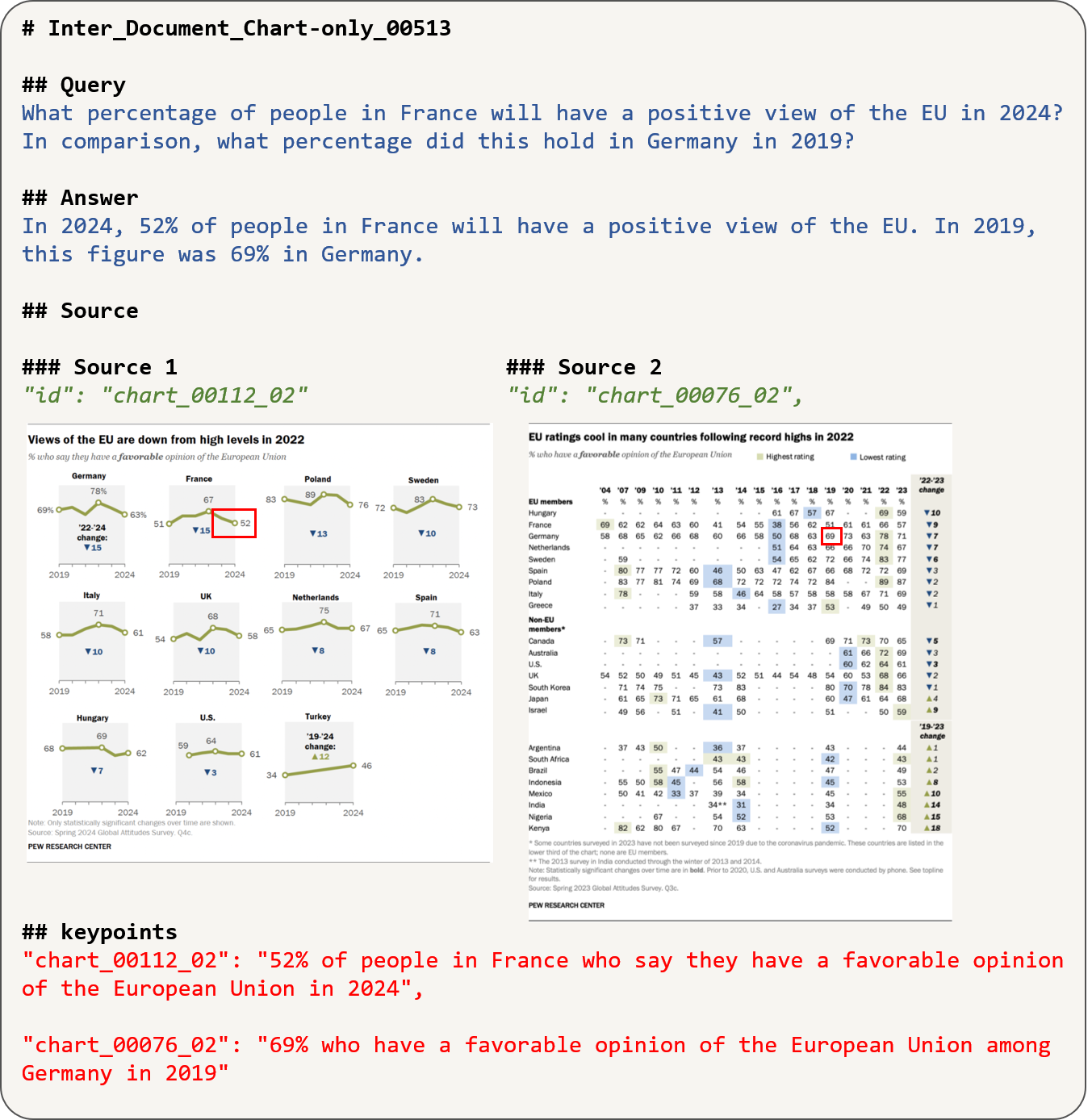}
  \caption{A sample case of inter-document chart-only question answering.}
  \label{sample_case_7}
\end{figure*}

\begin{figure*}[htbp] 
  \includegraphics[width=\textwidth]{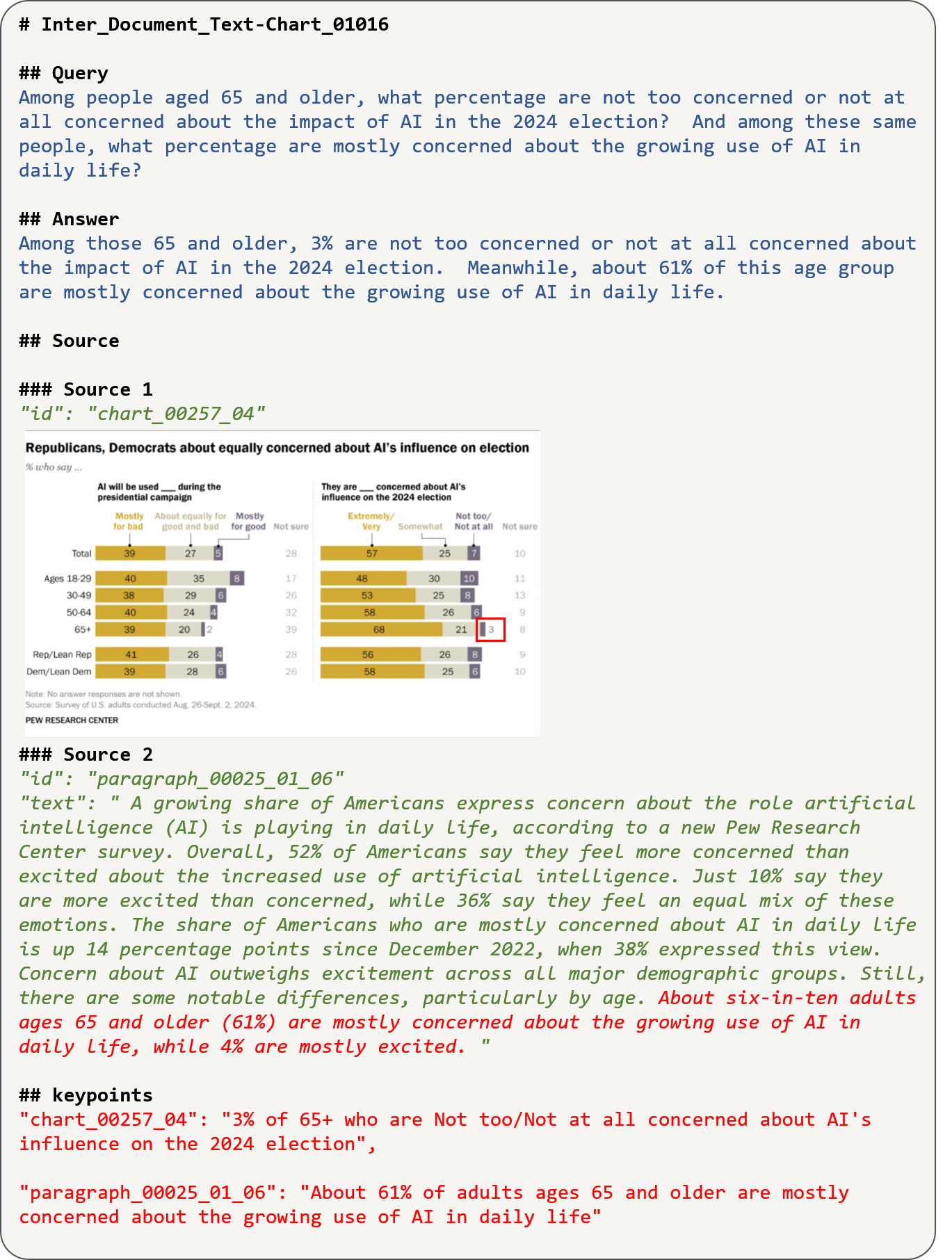}
  \caption{A sample case of inter-document text-chart question answering.}
  \label{sample_case_8}
\end{figure*}

We categorize the question-answering pairs in Chart-MRAG into eight distinct categories, as summarized in Table~\ref{tab:qa_taxonomy}, encompassing various combinations of single-point, intra-document, and inter-document scenarios across text-only, chart-only, and text-chart contexts. These categories are illustrated through representative examples: Single-Point Text-Only QA (Fig.~\ref{sample_case_1}), Single-Point Chart-Only QA (Fig.~\ref{sample_case_2}), Intra-Document Text-Only QA (Fig.~\ref{sample_case_3}), Intra-Document Chart-Only QA (Fig.~\ref{sample_case_4}), Intra-Document Text-Chart QA (Fig.~\ref{sample_case_5}), Inter-Document Text-Only QA (Fig.~\ref{sample_case_6}), Inter-Document Chart-Only QA (Fig.~\ref{sample_case_7}), and Inter-Document Text-Chart QA (Fig.~\ref{sample_case_8}).

\begin{table*}[htbp]
\centering
\setlength{\tabcolsep}{3pt}
\resizebox{\textwidth}{!}{
\small
\begin{tabular}{l|p{0.7\textwidth}}
\toprule
\multicolumn{2}{l}{\textbf{Source-Constrained and Modality-Constrained Question-Answer Categories}} \\
\midrule

\multirow{2}{*}{Single-Point Text-Only} & Questions that require reasoning about an individual textual keypoint ($k^t_i \in K^T$), focusing on discrete factual validation within a single text segment. \\
\midrule

\multirow{2}{*}{Single-Point Chart-Only} & Questions centered on an individual chart-only keypoint ($k^c_i \in K^C$), examining specific data points or visual elements within a single chart. \\
\midrule

\multirow{2}{*}{Intra-Document Text-Only} & Questions that necessitate integrative reasoning across multiple textual keypoints ($k^t_i, k^t_j \in K^T$) within the same document ($d_i \in D$). \\
\midrule

\multirow{2}{*}{Intra-Document Chart-Only} & Questions requiring comparative analysis of multiple chart-only keypoints ($k^c_i, k^c_j \in K^C$) from a single document ($d_i \in D$). \\
\midrule

\multirow{2}{*}{Intra-Document Text-Chart} & Questions involving cross-modal reasoning between textual and chart-only keypoints ($k^t_i \in K^T, k^c_j \in K^C$) within the same document ($d_i \in D$). \\
\midrule

\multirow{2}{*}{Inter-Document Text-Only} & Questions demanding associative reasoning between textual keypoints ($k^t_i, k^t_j \in K^T$) from distinct documents ($d_i, d_j \in D, i \neq j$). \\
\midrule

\multirow{2}{*}{Inter-Document Chart-Only} & Questions requiring comparative analysis of chart-only keypoints ($k^c_i, k^c_j \in K^C$) across different documents ($d_i, d_j \in D, i \neq j$). \\
\midrule

\multirow{2}{*}{Inter-Document Text-Chart} & Questions involving cross-modal and cross-document reasoning, integrating textual and chart keypoints ($k^t_i \in K^T, k^c_j \in K^C$) from different documents ($d_i, d_j \in D, i \neq j$). \\
\bottomrule
\end{tabular}
}
\caption{Taxonomy of question-answer pairs in Chart-MRAG, categorized by source constraints (Single-Point/Intra-Document/Inter-Document) and modality constraints (Text-only/Chart-only/Text-Chart).}
\label{tab:qa_taxonomy}
\end{table*}

\section{Retrieval Setup and Metrics}
\label{sec: retrieval_setup_and_metrics}

\subsection{Retrieval Setup}
For retrieval system, we designed three distinct configurations to evaluate different approaches to multimodal information retrieval:

\noindent \textbf{Unified Multimodal Embedding and Single Vector Store}. We directly embedded charts and text into a unified embedding space using vision-language models CLIP, JINA-CLIP, and SigLIP. This approach maps all content to same-dimensional vectors in a single vector store, enabling cross-modal matching between queries and documents regardless of their original modality.

\noindent \textbf{Multimodal Embeddings and Combined Vector Stores}. In this approach, charts are first converted to text summaries using \textit{GPT-4o}. Both these summaries and PDF-extracted text are then embedded using sparse BM25 and dense embedding models BGE-M3-base/large, E5-base/large into their respective vector stores. Similarity search in these embedding spaces retrieves relevant documents across both modalities.

\noindent \textbf{Multimodal Embeddings and Separate Vector Stores}. This approach maintains distinct embedding spaces for different modalities, leveraging specialized models for optimal representation. Charts are encoded using vision-language models (CLIP, JINA-CLIP, SigLIP), while textual content is processed through both sparse retrieval (BM25) and dense embedding models (BGE-M3-base/large, E5-base/large). The retrieval process operates in parallel across separate vector stores, with the final results aggregated using a weighted combination scheme.

\subsection{Retrieval Metrics}
We segment text into semantic chunks with an average length of 24.97 words, while treating each chart as an individual retrieval unit. We employ Recall@5 and Recall@10 as primary retrieval metrics. To ensure balanced representation, we implement a text-to-chart ratio of 3:2 in the final retrieval results.

Given that Chart-MRAG bench primarily consists of multi-hop questions requiring both textual and visual information, the comprehensive retrieval of all relevant sources is crucial for accurate answers. We employ $\text{Recall}@5$ and $\text{Recall}@10$ to evaluate the effectiveness and efficiency of the retrieval stage.

\textbf{Multimodal Recall.} We introduce a Multimodal RAG Retrieval Recall metric to evaluate the effectiveness of crossmodal retrieval process. For textual content, we perform sentence-level retrieval, while for charts, we treat each visualization as an individual reference unit. The Recall is formally defined as
\begin{equation}
    \text{Recall} = \frac{1}{n}\sum_{i=1}^n \mathbbm{1}(M(G_i,\mathcal{R})),
    \label{eq:recall}
\end{equation}

where $n$ is the total number of ground truth references (including both text chunks and charts), $G_i$ denotes the $i$-th ground truth reference, $\mathcal{R} = \{R_1, R_2, \ldots, R_k\}$ represents the set of retrieved references, $M(G_i,\mathcal{R})$ is a boolean function that returns true if (1) for textual references, all constituent sentences in $G_i$ are found in at least one reference in $\mathcal{R}$, or (2) for chart references, the exact chart is present in $\mathcal{R}$, and $\mathbbm{1}(\cdot)$ is the indicator function.

This metric assesses the crossmodal alignment between retrieved and ground truth references, where successful retrieval is determined by modality-specific criteria: sentence-level matching for text and exact matching for charts.

\section{Text-Over-Visual Modality Bias Case}
\label{sec: bias_case}

\begin{figure*}[htbp] 
  \includegraphics[width=\textwidth]{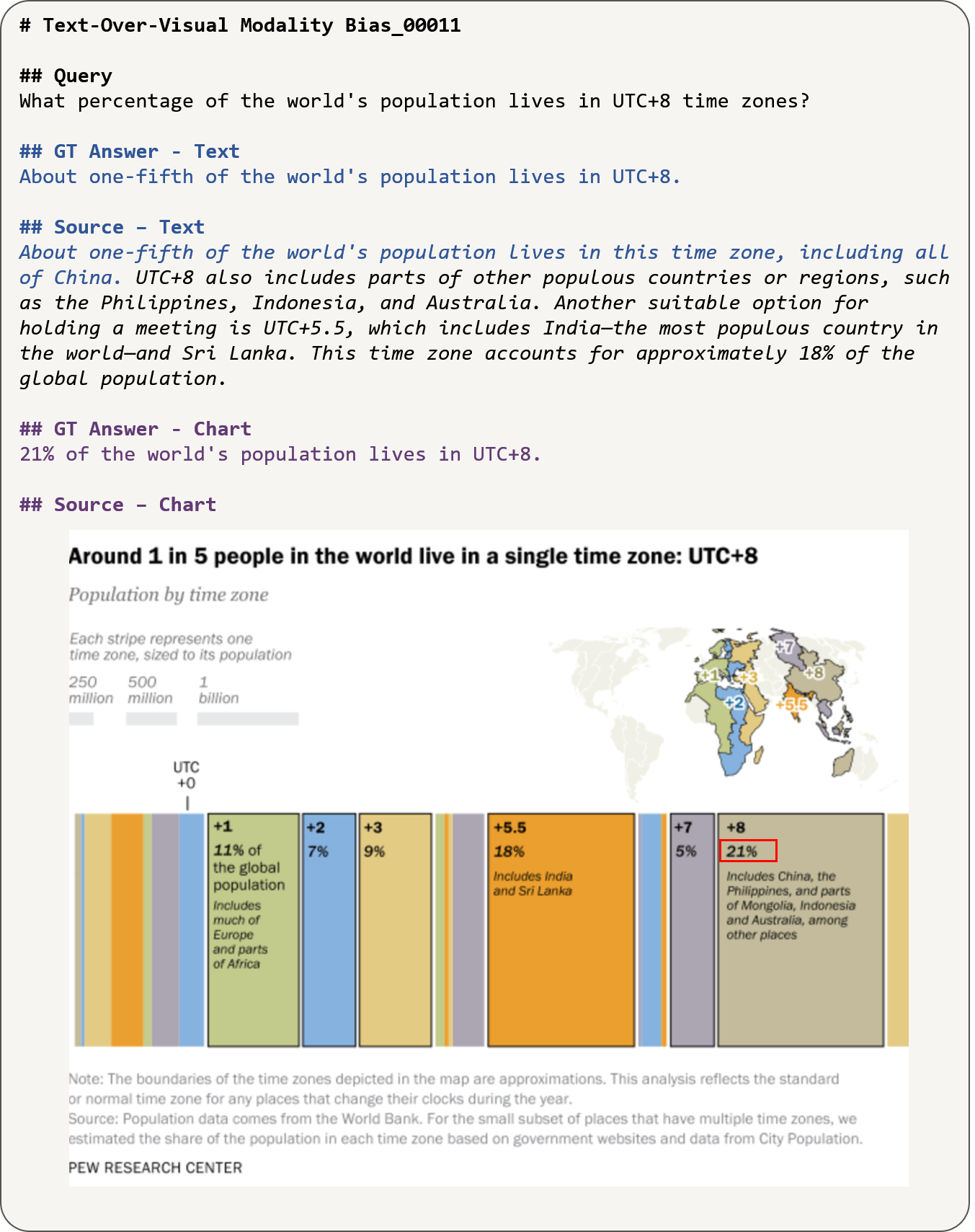}
  \caption{A Sample Case of Text-Over-Visual Modality Bias}
  \label{bias_case}
\end{figure*}

Section~\ref{bias_case} presents a comprehensive analysis of Text-Over-Visual Modality Bias, revealing a systematic preference for text-based processing across different model scales. Our experiments demonstrate that multimodal language models consistently favor text-only responses, even in scenarios where visual elements (particularly charts) contain more precise and relevant information. This bias raises important questions about the effective integration of multiple modalities in current AI systems.

Notably, our investigation reveals a clear correlation between model scale and the ability to handle multimodal information effectively. Larger MLLMs, particularly \textit{GPT-4o}, demonstrate sophisticated capabilities in detecting and managing information redundancy across modalities, proactively acknowledging such overlaps in 23\% of their responses. This behavior suggests a more nuanced understanding of the complementary nature of different information sources.

In contrast, smaller models exhibit significant limitations in processing multimodal inputs. For instance, SAIL-VL-2B (2B parameters) shows a stark inability to integrate information across modalities, highlighting the critical role of model scale in achieving effective multimodal reasoning.

\section{Model Prompts}
\label{sec: prompts}
\begin{figure*}[htbp] 
  \includegraphics[width=\textwidth]{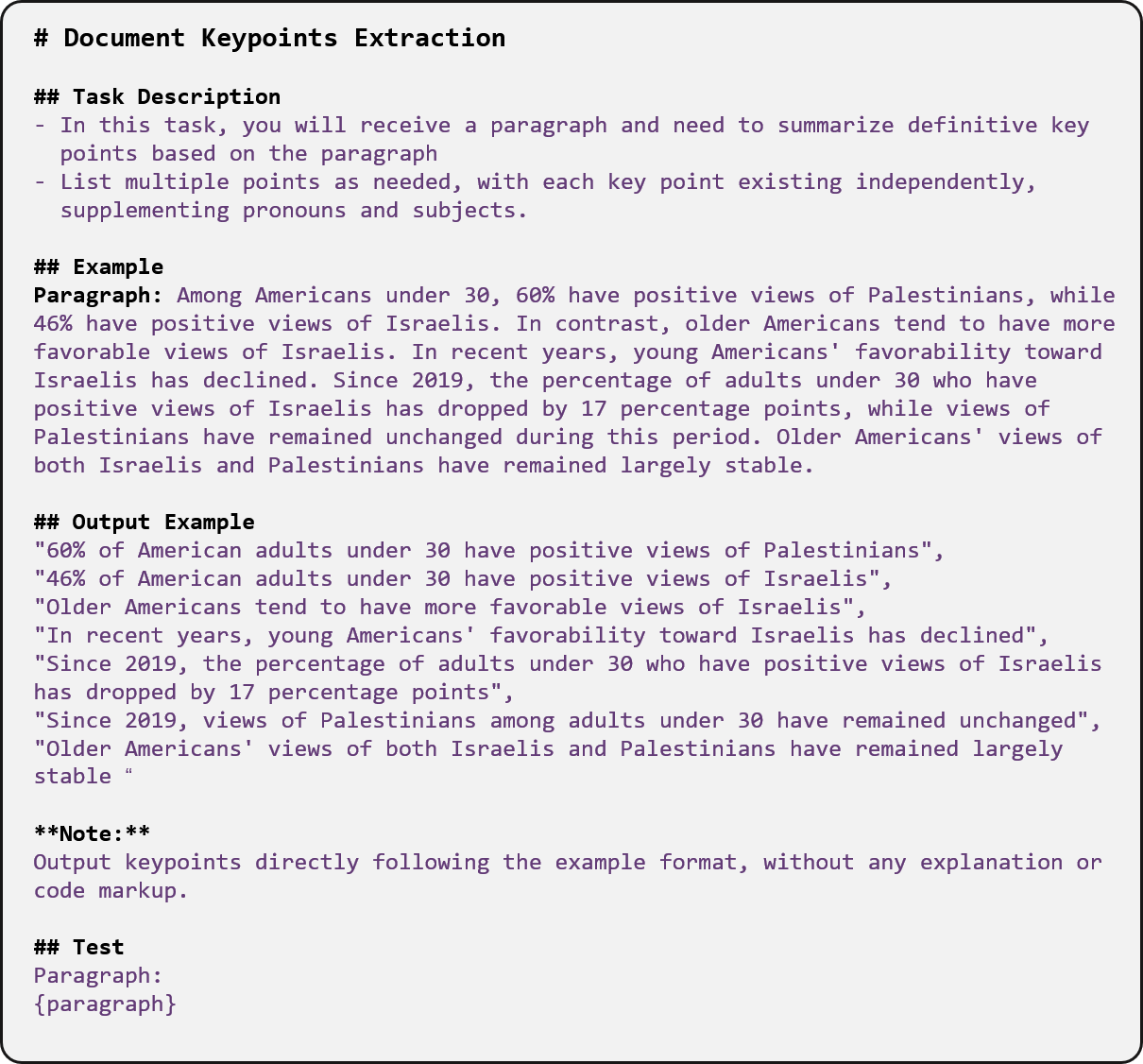}
  \caption{Document keypoints extraction prompt details.}
  \label{prompt1}
\end{figure*}

\begin{figure*}[htbp] 
  \includegraphics[width=\textwidth]{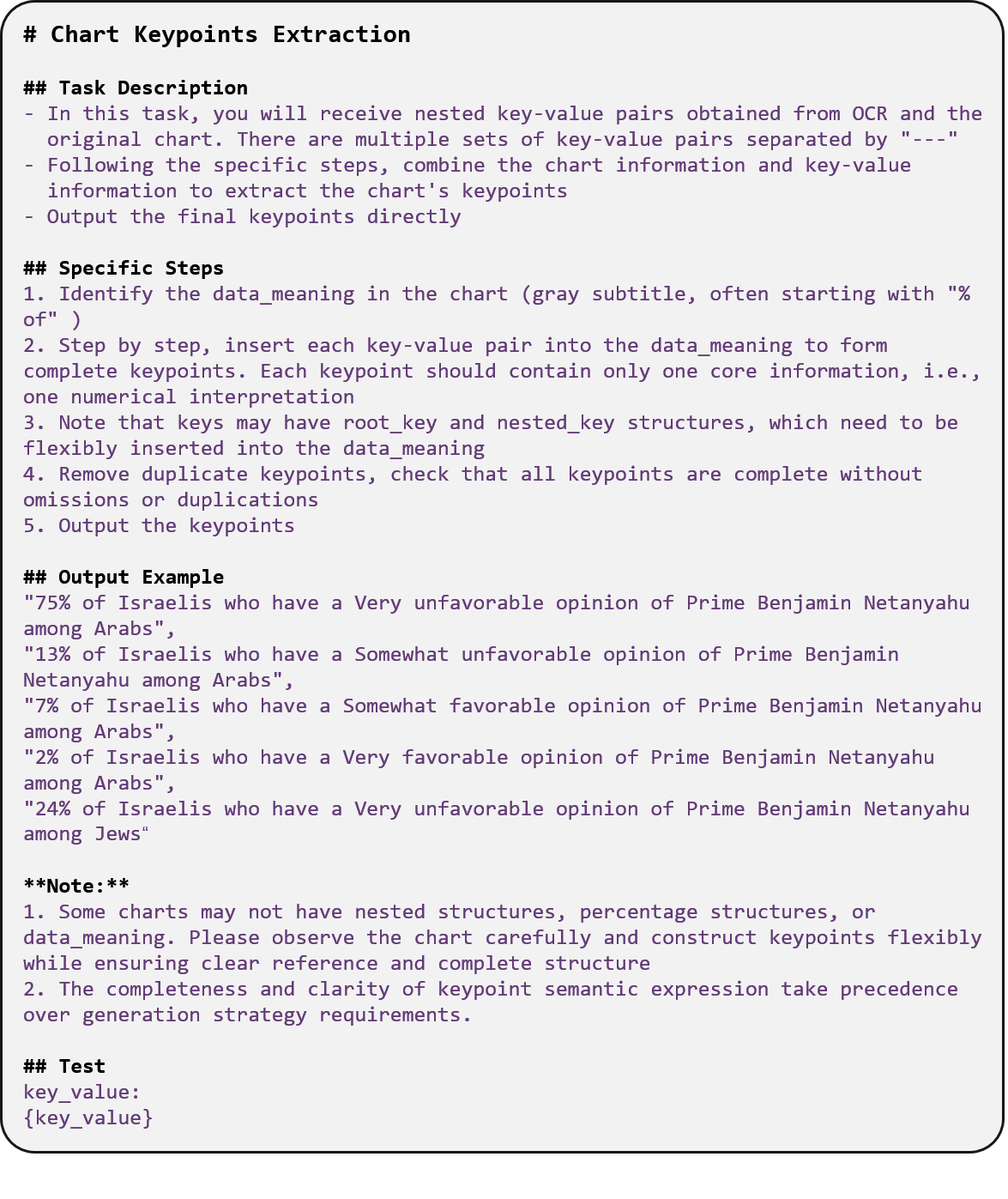}
  \caption{Chart keypoints extraction prompt details.}
  \label{prompt2}
\end{figure*}

CHARGE framework encompasses multiple stages, each guided by specific prompts designed to facilitate different aspects of the process. We detail these prompts according to their respective stages:

In the Extract Keypoints stage, we employ two specialized prompts: one for document keypoint extraction (Fig.~\ref{prompt1}) and another for chart keypoint extraction (Fig.~\ref{prompt2}). These prompts are designed to identify and extract crucial information points from both textual and visual components.

\begin{figure*}[htbp] 
  \includegraphics[width=\textwidth]{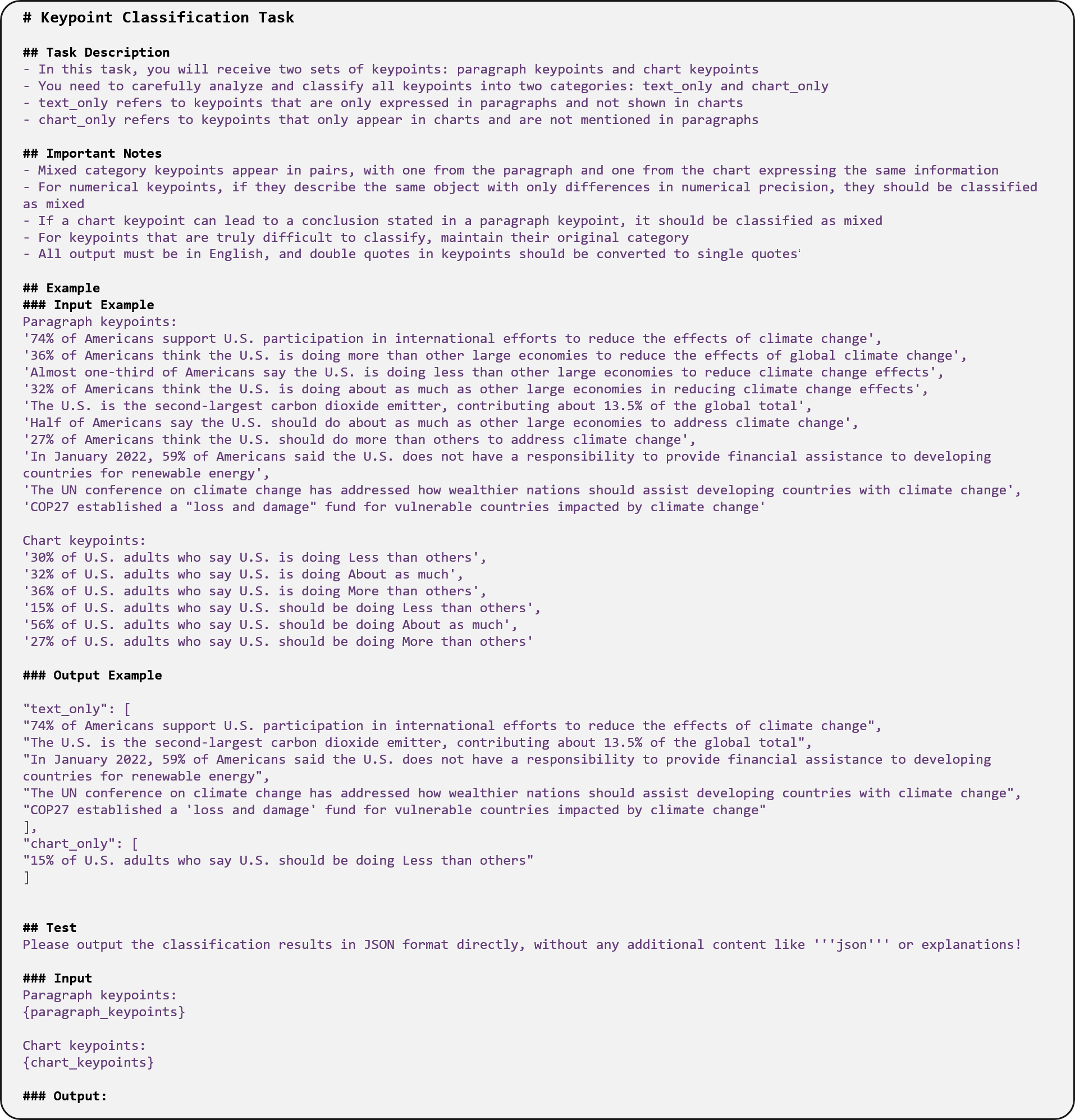}
  \caption{Keypoint classification task prompt details.}
  \label{prompt3}
\end{figure*}

\begin{figure*}[htbp]
  \includegraphics[width=\textwidth]{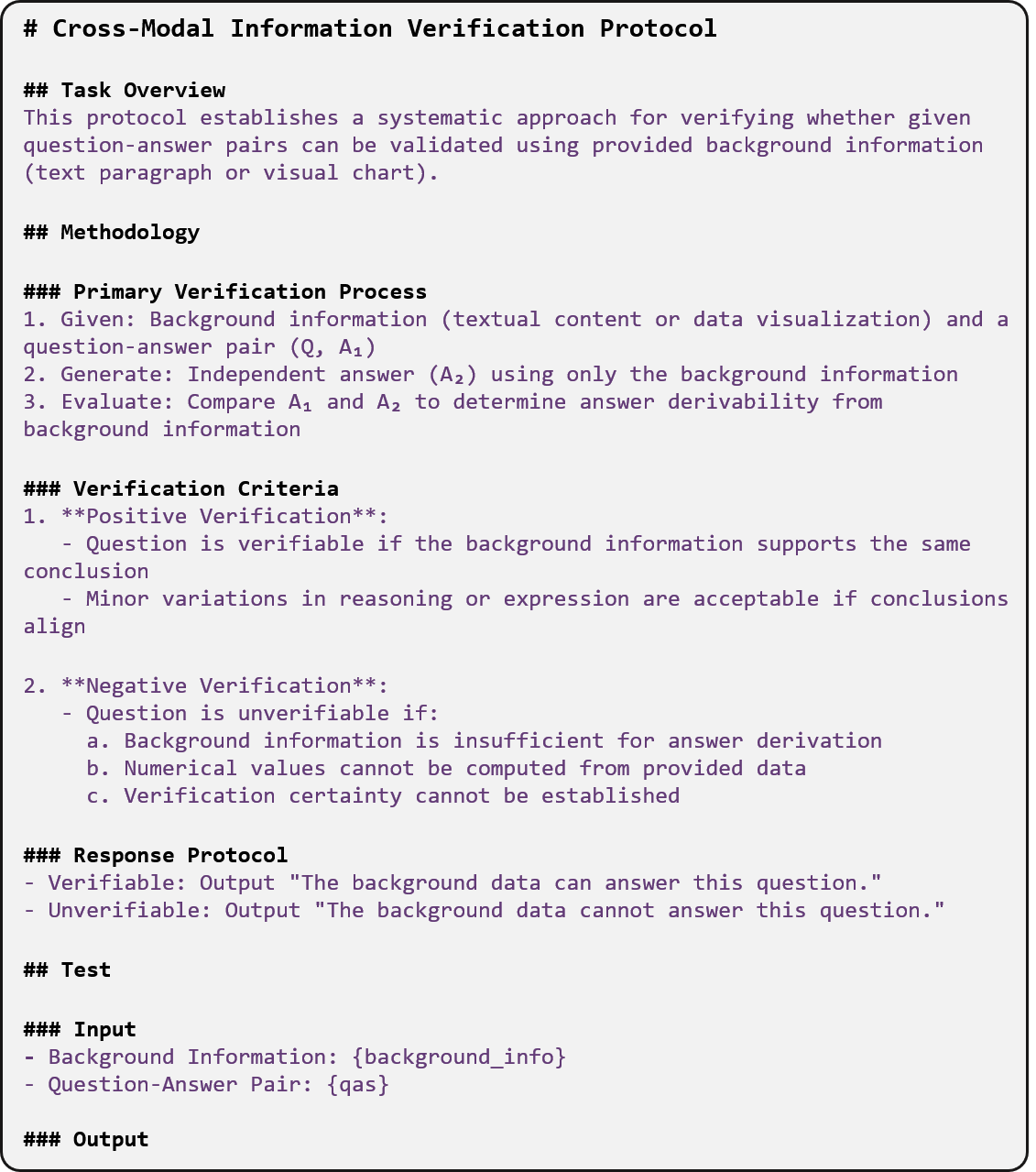}
  \caption{Cross-modal information verification protocol prompt details.}
  \label{prompt4}
\end{figure*}

The Crossmodal Verification stage utilizes two key prompts: a keypoint classification prompt (Fig.~\ref{prompt3}) and a cross-modal information verification protocol (Fig.~\ref{prompt4}). These prompts work in tandem to ensure the consistency and accuracy of information across different modalities.

\begin{figure*}[htbp]
\includegraphics[width=\textwidth]{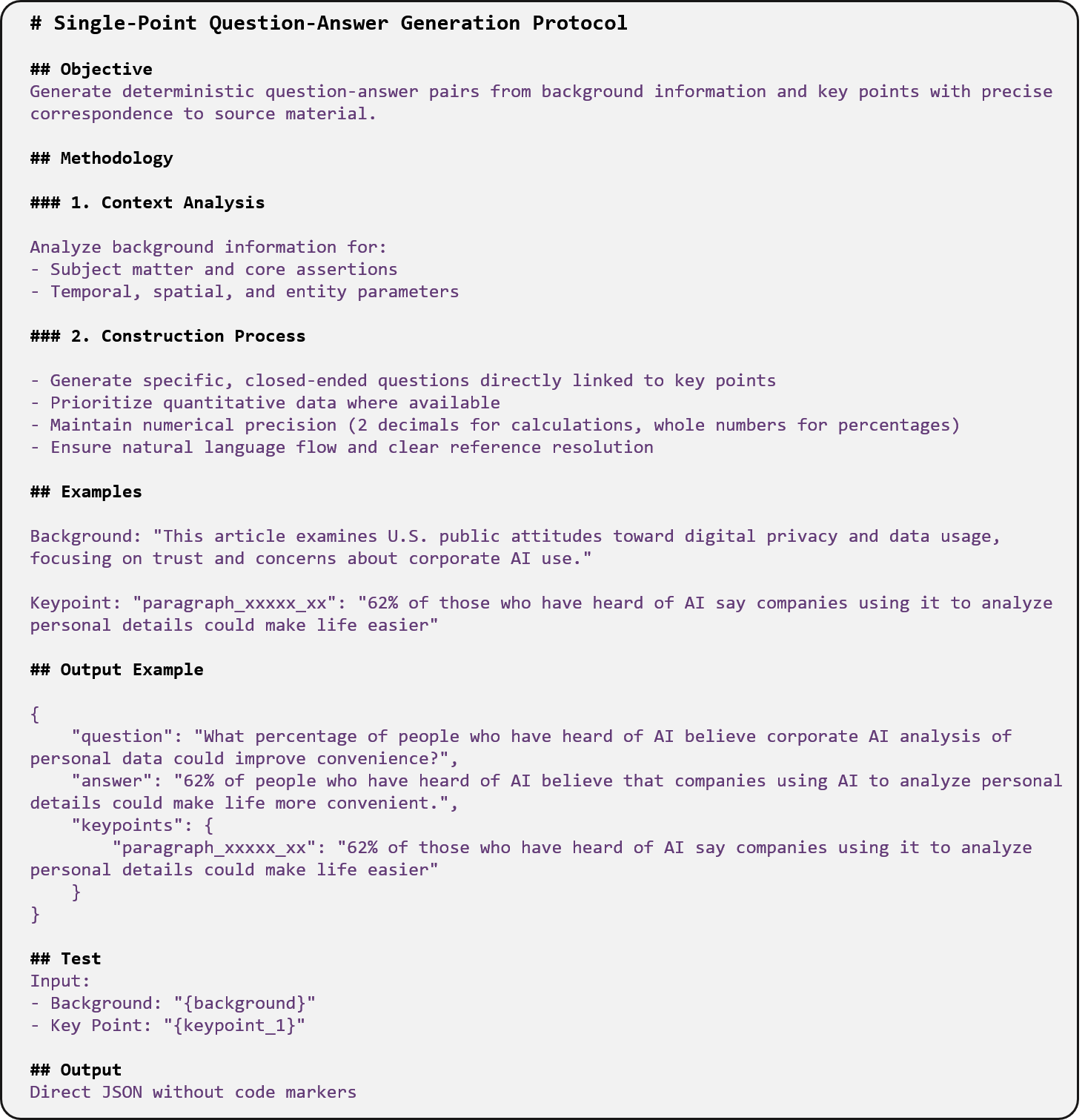}
\caption{Single-point question-answer generation protocol prompt details.}
\label{prompt5}
\end{figure*}

\begin{figure*}[htbp] 
  \includegraphics[width=\textwidth]{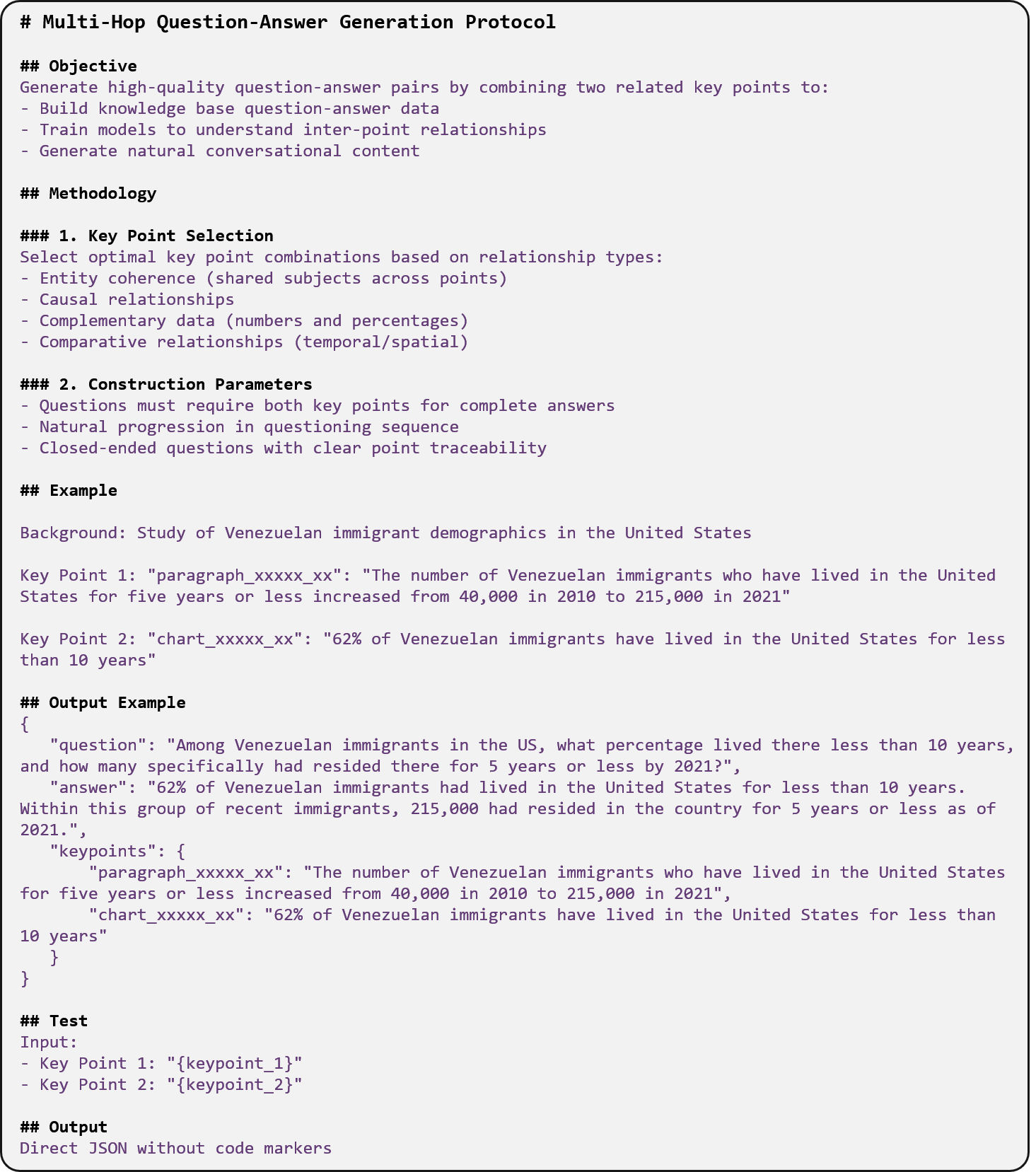}
  \caption{Multi-hop question-answer generation protocol prompt details.}
  \label{prompt6}
\end{figure*}

For Question-Answer Pair Generation, we implement two distinct protocols: a single-point generation protocol (Fig.~\ref{prompt5}) for straightforward questions, and a multi-hop generation protocol (Fig.~\ref{prompt6}) for complex questions requiring multiple reasoning steps.

\begin{figure*}[htbp] 
  \includegraphics[width=\textwidth]{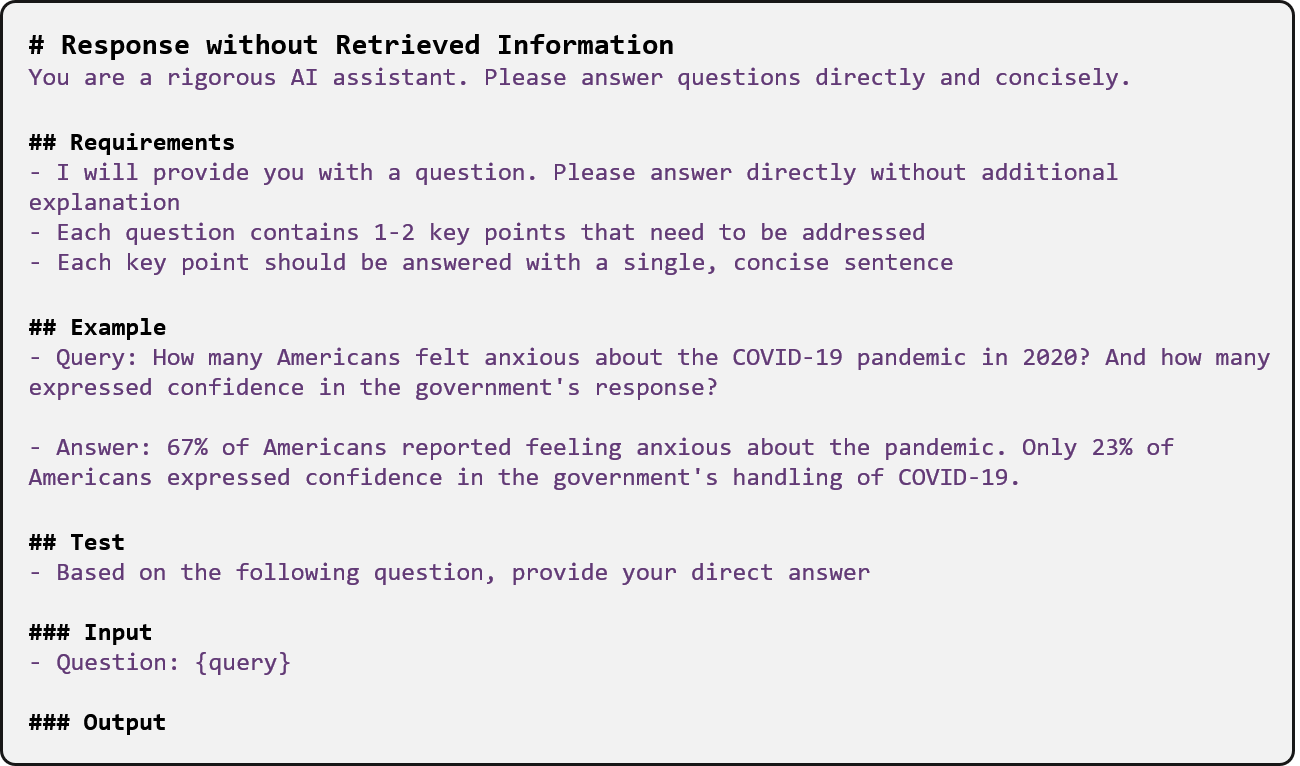}
  \caption{Response without retrieved information prompt details.}
  \label{prompt7}
\end{figure*}

\begin{figure*}[htbp] 
  \includegraphics[width=\textwidth]{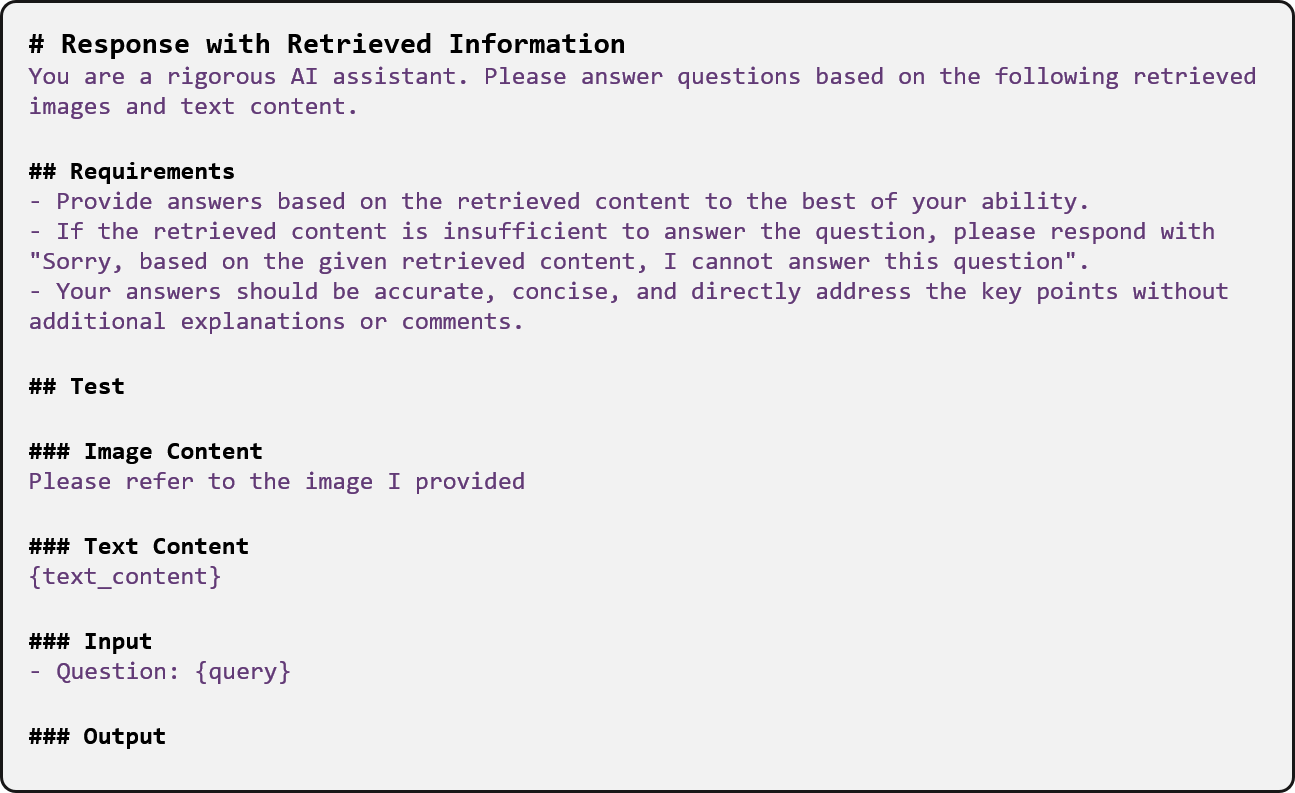}
  \caption{Response with retrieved information prompt details.}
  \label{prompt8}
\end{figure*}

The Response stage features two prompts: one designed for generating responses without retrieved information (Fig.~\ref{prompt7}), and another for responses incorporating retrieved information (Fig.~\ref{prompt8}). This dual approach enables flexible response generation based on available context.

\begin{figure*}[htbp] 
  \includegraphics[width=\textwidth]{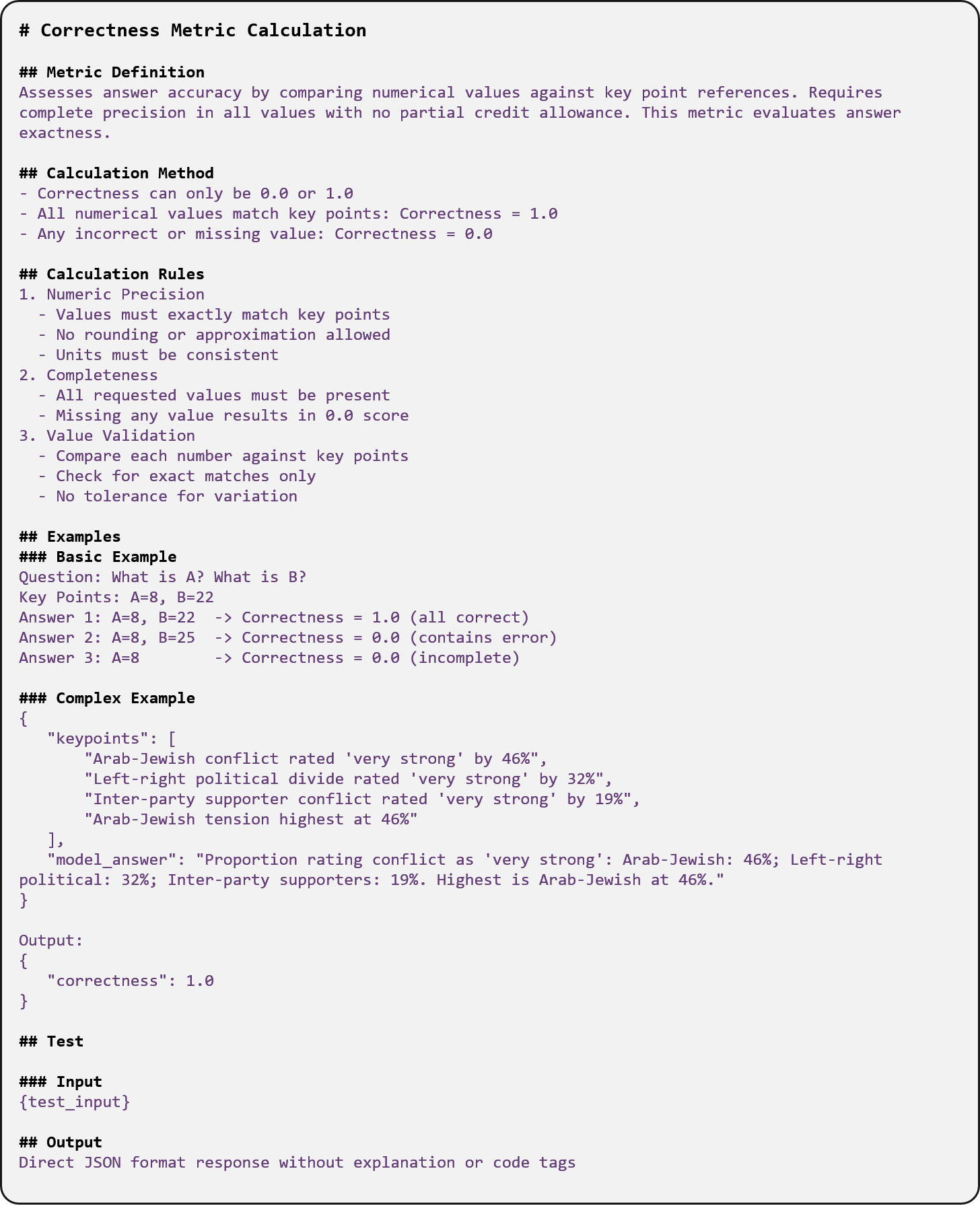}
  \caption{Correctness metric calculation prompt details.}
  \label{prompt9}
\end{figure*}

\begin{figure*}[htbp] 
  \includegraphics[width=\textwidth]{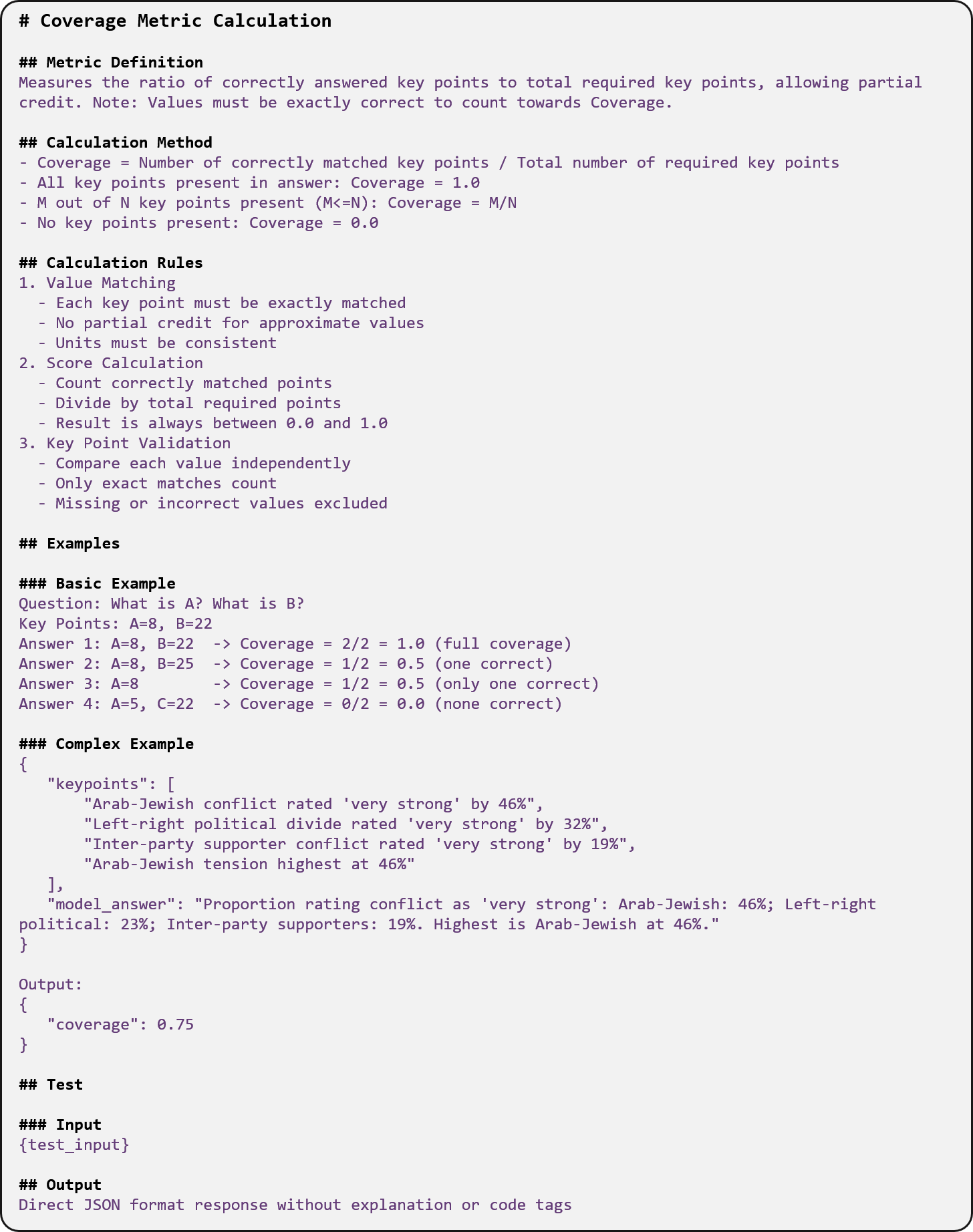}
  \caption{Coverage metric calculation prompt details.}
  \label{prompt10}
\end{figure*}

Finally, the Evaluation stage employs two metric calculation prompts: one for assessing correctness (Fig.~\ref{prompt9}) and another for measuring coverage (Fig.~\ref{prompt10}). These prompts ensure comprehensive evaluation of the generated responses.

\label{sec:chartmrag_cases_1}

\end{document}

%% file: retrieved_tables.tex
\begin{table*}[t]
\centering
\setlength{\tabcolsep}{3pt}  
\resizebox{\textwidth}{!}{  
\begin{tabular}{l|cc|cccc|cccccc|cccccc}
\toprule
\multirow{3}{*}{Model} & \multicolumn{2}{c|}{Overall} & \multicolumn{4}{c|}{Single-Point} & \multicolumn{6}{c|}{Intra-Document} & \multicolumn{6}{c}{Inter-Document} \\
\cmidrule{2-19}
& \multirow{2}{*}{R@5} & \multirow{2}{*}{R@10} & \multicolumn{2}{c}{Text-only} & \multicolumn{2}{c|}{Chart-only} & \multicolumn{2}{c}{Text-only} & \multicolumn{2}{c}{Chart-only} & \multicolumn{2}{c|}{Text-Chart} & \multicolumn{2}{c}{Text-only} & \multicolumn{2}{c}{Chart-only} & \multicolumn{2}{c}{Text-Chart} \\
\cmidrule{4-19}
& & & R@5 & R@10 & R@5 & R@10 & R@5 & R@10 & R@5 & R@10 & R@5 & R@10 & R@5 & R@10 & R@5 & R@10 & R@5 & R@10 \\
\midrule
\multicolumn{19}{l}{\textit{\textbf{Method 1: Unified Multimodal Embedding and Single Vector Store}}} \\
\midrule
SigLIP & 11.69 & 15.95 & 50.00 & 57.07 & 0.00 & 0.00 & 19.63 & 30.74 & 0.00 & 0.00 & 0.00 & 0.00 & 16.20 & 27.31 & 0.00 & 0.00 & 0.00 & 0.00 \\
CLIP & 13.26 & 19.07 & 56.06 & 65.15 & 0.00 & 0.00 & 24.44 & 42.22 & 0.00 & 0.00 & 0.00 & 0.00 & 16.20 & 28.70 & 0.00 & 0.00 & 0.00 & 0.00 \\
JINA & 23.14 & 29.02 & {\colorbox{lightblue!75}{77.78}} & {\colorbox{lightblue!75}{85.35}} & 0.00 & 0.00 & {\colorbox{lightblue!75}{47.04}} & {\colorbox{lightblue!75}{63.70}} & 0.00 & 0.00 & 0.00 & 0.00 & {\colorbox{lightblue!75}{41.20}} & {\colorbox{lightblue!75}{56.94}} & 0.00 & 0.00 & 0.00 & 0.00 \\
\midrule
\multicolumn{19}{l}{\textit{\textbf{Method 2: Multimodal Embeddings and Combined Vector Stores (Caption generated by GPT-4-Vision)}}} \\
\midrule
BGE-M3-base & 22.89 & 31.21 & 39.90 & 47.47 & 52.24 & 62.09 & 9.63 & 18.52 & 17.01 & 31.29 & 9.09 & 15.51 & 9.72 & 15.28 & 13.43 & 19.40 & 4.46 & 11.61 \\
BM25 & 27.02 & 36.46 & 51.01 & 54.55 & 52.24 & 63.88 & 16.67 & 26.67 & 12.93 & 25.17 & 7.49 & 19.79 & 23.15 & 31.94 & 10.45 & 16.42 & 12.50 & 21.43 \\
BGE-M3-large & 27.64 & 39.52 & 64.65 & 70.71 & 43.58 & 59.70 & 29.26 & 42.96 & 10.20 & 17.69 & 8.02 & 18.72 & 18.98 & 32.41 & 5.97 & 14.93 & 8.93 & 22.32 \\
E5-base & 35.27 & 47.59 & 67.17 & 73.74 & 66.27 & {80.90} & 21.48 & 34.81 & 23.13 & 47.62 & 15.51 & 25.13 & 20.37 & 27.78 & {\colorbox{lightblue!75}{20.90}} & 35.07 & 14.29 & 23.21 \\
E5-large & {41.53} & {59.54} & {72.73} & {79.80} & 64.78 & 79.40 & {38.89} & {60.74} & 23.13 & 48.30 & {18.18} & {\colorbox{lightblue!75}{41.71}} & {35.65} & {53.24} & {\colorbox{lightblue!75}{20.90}} & {41.04} & {\colorbox{lightblue!75}{22.32}} & {\colorbox{lightblue!75}{40.18}} \\
\midrule
\multicolumn{19}{l}{\textit{\textbf{Method 3: Multimodal Embeddings and Separate Vector Stores}}} \\
\midrule
JINA + BM25 & 23.83 & 33.90 & 48.48 & 53.03 & 45.67 & 59.10 & 11.85 & 20.37 & 14.97 & 28.57 & 9.63 & 18.72 & 15.74 & 26.39 & 7.46 & 19.40 & 14.29 & 21.43 \\
CLIP + BGE-M3-base & 25.64 & 36.09 & 34.34 & 41.92 & {66.57} & 77.61 & 5.93 & 12.96 & {24.49} & {51.70} & 10.70 & 19.25 & 6.48 & 12.04 & {14.93} & 35.07 & 11.61 & 12.50 \\
CLIP + BGE-M3-large  & 33.40 & 46.97 & 57.58 & 68.18 & {66.57} & 77.61 & 20.74 & 34.07 & {24.49} & {51.70} & 19.79 & {32.09} & 13.89 & 24.07 & {14.93} & 35.07 & {16.07} & 25.89 \\
SigLIP + E5-base & 37.96 & 52.47 & 64.65 & 69.70 & {\colorbox{lightblue!75}{84.18}} & {\colorbox{lightblue!75}{93.73}} & 15.56 & 29.63 & {\colorbox{lightblue!75}{39.46}} & {\colorbox{lightblue!75}{74.15}} & 17.11 & 28.34 & 13.89 & 24.54 & 14.18 & {\colorbox{lightblue!75}{44.78}} & 14.29 & {28.57} \\
SigLIP + E5-large & {\colorbox{lightblue!75}{42.53}} & {\colorbox{lightblue!75}{61.10}} & 68.69 & 75.76 & {\colorbox{lightblue!75}{84.18}} & {\colorbox{lightblue!75}{93.73}} & 25.19 & 47.41 & {\colorbox{lightblue!75}{39.46}} & {\colorbox{lightblue!75}{74.15}} & {\colorbox{lightblue!75}{24.06}} & {\colorbox{lightblue!75}{41.71}} & 21.76 & 43.06 & 14.18 & {\colorbox{lightblue!75}{44.78}} & {\colorbox{lightblue!75}{22.32}} & {\colorbox{lightblue!75}{40.18}} \\
\bottomrule
\end{tabular}
}
\caption{Performance Comparison of Different Multimodal Retrieved Models (\%) on Chart-MRAG benchmark, evaluating three strategies: Unified Multimodal Embedding with Single Vector Store, Multimodal Embeddings with Combined Vector Stores, and Multimodal Embeddings with Separate Vector Stores (best scores highlighted in blue).}
\label{tab:retrieved_performance}
\end{table*}

%% file: generation_table.tex
\begin{table*}[t]
\centering
\setlength{\tabcolsep}{3pt}  
\resizebox{\textwidth}{!}{  
\begin{tabular}{l|cc|c|c|cc|cc|cc|cc|cc|cc}
\toprule
\multirow{3}{*}{Model} & \multicolumn{2}{c|}{Overall} & \multicolumn{2}{c|}{Single-Point} & \multicolumn{6}{c|}{Intra-Document} & \multicolumn{6}{c}{Inter-Document} \\
\cmidrule{2-17}
& \multirow{2}{*}{Corr.} & \multirow{2}{*}{Cov.} & \multicolumn{1}{c}{Text-only} & \multicolumn{1}{c|}{Chart-only} & \multicolumn{2}{c}{Text-only} & \multicolumn{2}{c}{Chart-only} & \multicolumn{2}{c|}{Text-Chart} & \multicolumn{2}{c}{Text-only} & \multicolumn{2}{c}{Chart-only} & \multicolumn{2}{c}{Text-Chart} \\
\cmidrule{4-17}
& &  & Corr. & Corr. & Corr. & Cov. & Corr. & Cov. & Corr. & Cov. & Corr. & Cov. & Corr. & Cov. & Corr. & Cov. \\
\midrule
\multicolumn{17}{l}{\textit{\textbf{Open-Source VLMs}}} \\
\midrule
SAIL-VL-2B & 0.38 & 1.58 & 1.52 &	0.30 &	\colorbox{lightred!75}{0.74} &	2.59 &	\colorbox{lightred!75}{0.00} &	0.00 &	\colorbox{lightred!75}{0.00} &	1.60 &	0.00 &	3.47 &	\colorbox{lightred!75}{0.00} &	0.37 &	\colorbox{lightred!75}{0.00} &	1.79 \\
\hspace{2em}  + RAG\,\textit{(k=5)} & 3.88 &	8.51 &	19.19   &	4.18  &	1.48  &	14.44  &	0.00  &	2.04 &	0.00  &	1.34  &	2.78  &	13.89 &	0.00  &	0.75  &	0.00  &	1.79 \\
\hspace{2em}  + RAG\,\textit{(k=10)} & 3.19 & 7.71 & 14.65 & 4.48 & 2.22 & 13.89 & 0.00 & 1.36 & 0.00 & 2.14 & 0.46 & 11.11 & 0.00 & 0.75 & 0.00 & 3.12 \\
\hspace{2em}  + RAG\,\textit{(GT)} & 19.82 & 29.44 & 63.64 & 9.85 & 31.30 & 57.22 & 2.72 & 6.80 & 0.53 & 5.61 & 29.86 & 54.17 & 0.00 & 3.36 & 3.57 & 12.05 \\

\midrule
qwen2-VL-7B-instruct & 1.16 & \colorbox{lightred!75}{4.45} & 4.55 & 2.09 & 0.56 & 7.22 & \colorbox{lightred!75}{0.00} & 1.19 & \colorbox{lightred!75}{0.00} & 3.48 & \colorbox{lightred!75}{0.46} & \colorbox{lightred!75}{7.41} & \colorbox{lightred!75}{0.00} & 1.49 & \colorbox{lightred!75}{0.00} & \colorbox{lightred!75}{5.36} \\

\hspace{2em}  + RAG\,\textit{(k=5)} & 13.51 & 23.55 & \colorbox{lightred!75}{50.51} & 4.78 & \colorbox{lightred!75}{22.41} & 46.73 & 1.36 & 3.40 & 1.07 & 8.82 & 15.51 & 41.67 & \colorbox{lightred!75}{1.49} & 2.99 & 0.00 & 9.82 \\

\hspace{2em}  + RAG\,\textit{(k=10)} & 14.45 & 23.57 & 51.01 & 3.28 & 26.11 & 49.81 & 0.68 & 2.38 & 1.60 & 7.75 & 20.14 & 43.75 & 0.75 & 1.49 & 0.00 & 8.48 \\

\hspace{2em}  + RAG\,\textit{(GT)} & 33.15 & 42.46 & 78.28 & 11.04 & 64.26 & 81.30 & 2.04 & 9.52 & 5.88 & 20.86 & 62.73 & 80.56 & 2.99 & 6.72 & 9.82 & 27.23 \\

\midrule
MiniCPM-V-2.6-8B & 0.88 & 4.05 & 2.02 & \colorbox{lightred!75}{2.69} & 0.37 & 6.67 & \colorbox{lightred!75}{0.00} & \colorbox{lightred!75}{1.70} & \colorbox{lightred!75}{0.00} & \colorbox{lightred!75}{3.74} & 0.00 & 6.71 & \colorbox{lightred!75}{0.00} & 2.24 & \colorbox{lightred!75}{0.00} & 4.46 \\

\hspace{2em}  + RAG\,\textit{(k=5)} & 17.32 & 31.32 & 47.98 & 25.67 & 14.81 & 42.59 & 3.40 & 13.61 & 4.01 & \colorbox{lightred!75}{20.59} & 15.97 & 43.52 & \colorbox{lightred!75}{1.49} & 11.07 & 6.25 & 22.77 \\

\hspace{2em}  + RAG\,\textit{(k=10)} & 17.60 & 31.51 & 48.48 & 19.70 & 21.11 & 50.43 & 2.72 & 12.24 & 4.81 & 19.79 & 19.68 & 46.30 & 1.49 & 11.07 & 4.46 & 23.21 \\

\hspace{2em}  + RAG\,\textit{(GT)} & 46.94 & 59.41 & 79.29 & 48.66 & \colorbox{lightred!75}{65.37} & 80.99 & 12.93 & 29.93 & \colorbox{lightred!75}{22.46} & 46.79 & \colorbox{lightred!75}{69.68} & \colorbox{lightred!75}{83.10} & 14.18 & 32.34 & 20.98 & 47.32 \\

\midrule
Llama-3.2-90B-Vision & \colorbox{lightred!75}{1.22} & 4.36 & \colorbox{lightred!75}{5.56} & 2.09 & 0.37 & \colorbox{lightred!75}{8.15} & \colorbox{lightred!75}{0.00} & 1.36 & \colorbox{lightred!75}{0.00} & 1.87 & 0.23 & 5.79 & \colorbox{lightred!75}{0.00} & \colorbox{lightred!75}{3.36} & \colorbox{lightred!75}{0.00} & 4.46 \\

\hspace{2em}  + RAG\,\textit{(k=5)} & \colorbox{lightred!75}{20.42} & \colorbox{lightred!75}{34.68} & \colorbox{lightred!75}{50.51} & \colorbox{lightred!75}{30.15} & 21.85 & \colorbox{lightred!75}{49.20} & \colorbox{lightred!75}{5.44} & \colorbox{lightred!75}{16.21} & \colorbox{lightred!75}{4.81} & 20.05 & \colorbox{lightred!75}{17.82} & \colorbox{lightred!75}{45.83} & \colorbox{lightred!75}{1.49} & \colorbox{lightred!75}{12.69} & \colorbox{lightred!75}{8.04} & \colorbox{lightred!75}{28.57} \\

\hspace{2em}  + RAG\,\textit{(k=10)} & \colorbox{lightred!75}{23.11} & \colorbox{lightred!75}{37.31} & \colorbox{lightred!75}{53.54} & \colorbox{lightred!75}{31.94} & \colorbox{lightred!75}{26.67} & \colorbox{lightred!75}{53.15} & \colorbox{lightred!75}{4.76} & \colorbox{lightred!75}{16.67} & \colorbox{lightred!75}{5.88} & \colorbox{lightred!75}{24.33} & \colorbox{lightred!75}{23.38} & \colorbox{lightred!75}{49.54} & \colorbox{lightred!75}{4.48} & \colorbox{lightred!75}{15.67} & \colorbox{lightred!75}{8.93} & \colorbox{lightred!75}{30.36} \\

\hspace{2em}  + RAG\,\textit{(GT)} & \colorbox{lightred!75}{50.16} & \colorbox{lightred!75}{64.03} & \colorbox{lightred!75}{79.80} & \colorbox{lightred!75}{58.81} & 63.33 & \colorbox{lightred!75}{81.36} & \colorbox{lightred!75}{21.09} & \colorbox{lightred!75}{40.95} & 21.66 & \colorbox{lightred!75}{48.13} & 59.49 & 78.24 & \colorbox{lightred!75}{32.09} & \colorbox{lightred!75}{46.27} & \colorbox{lightred!75}{29.46} & \colorbox{lightred!75}{57.59} \\

\midrule
\multicolumn{17}{l}{\textit{\textbf{Proprietary VLMs}}} \\
\midrule
GPT-4o & 2.50 & 8.32 & \colorbox{lightblue!75}{8.59} & \colorbox{lightblue!75}{5.97} & 0.37 & 13.64 & \colorbox{lightblue!75}{0.00} & 2.38 & 0.00 & 5.08 & 0.46 & 12.27 & 0.75 & 2.61 & 0.00 & 8.48 \\

\hspace{2em}  + RAG\,\textit{(k=5)} & 25.89 & 36.63 & 47.98 & 45.97 & \colorbox{lightblue!75}{24.44} & 43.89 & 11.90 & 20.75 & \colorbox{lightblue!75}{8.29} & 22.99 & 19.44 & 37.96 & 6.72 & 13.06 & 13.39 & 30.36 \\

\hspace{2em}  + RAG\,\textit{(k=10)} & 29.11 & 41.07 & 47.98 & 47.76 & 28.52 & 51.85 & 13.27 & 23.47 & \colorbox{lightblue!75}{10.96} & 26.74 & \colorbox{lightblue!75}{25.93} & 43.98 & 14.93 & 23.88 & 15.62 & 34.38 \\

\hspace{2em}  + RAG\,\textit{(GT)} & 52.25 & 62.97 & 70.20 & 63.88 & 64.07 & 77.22 & 24.83 & 38.10 & 22.19 & 45.19 & 62.96 & 75.46 & 36.57 & 51.12 & \colorbox{lightblue!75}{41.52} & 62.95 \\

\midrule
GPT-4-Vision & 2.13 & 7.82 & 6.57 & 4.78 & 0.74 & 11.91 & \colorbox{lightblue!75}{0.00} & 3.06 & 0.00 & 5.61 & 1.39 & \colorbox{lightblue!75}{15.51} & 0.00 & 2.99 & 0.00 & 6.25 \\

\hspace{2em}  + RAG\,\textit{(k=5)} & 25.17 & 35.96 & 46.97 & 45.37 & 21.85 & 42.04 & 12.93 & 22.45 & 6.95 & 22.46 & 21.06 & 38.66 & 7.46 & 14.55 & 9.82 & 26.34 \\

\hspace{2em}  + RAG\,\textit{(k=10)} & 29.21 & 41.21 & 49.49 & 48.06 & 31.11 & 51.20 & 11.56 & 23.47 & 10.43 & 27.01 & 25.46 & 46.30 & 12.69 & 23.13 & 13.84 & 33.93 \\

\hspace{2em}  + RAG\,\textit{(GT)} & 53.85 & 63.95 & 71.72 & 64.78 & 67.78 & 79.91 & 25.17 & 39.68 & \colorbox{lightblue!75}{25.67} & 47.33 & 67.13 & 78.01 & 38.06 & 50.37 & 33.93 & 56.25 \\

\midrule
Gemini-1.5-Pro & 2.47 & 8.11 & 7.58 & 3.58 & \colorbox{lightblue!75}{2.04} & \colorbox{lightblue!75}{15.25} & \colorbox{lightblue!75}{0.00} & 2.72 & 0.00 & 4.55 & 1.85 & 13.66 & \colorbox{lightblue!75}{1.49} & 5.97 & \colorbox{lightblue!75}{0.89} & 4.91 \\

\hspace{2em}  + RAG\,\textit{(k=5)} & 26.64 & 43.63 & \colorbox{lightblue!75}{51.52} & 45.97 & 24.07 & 52.04 & 11.56 & 30.95 & 7.75 & 31.82 & \colorbox{lightblue!75}{21.53} & 46.76 & \colorbox{lightblue!75}{8.21} & 27.74 & 14.29 & 38.39 \\

\hspace{2em}  + RAG\,\textit{(k=10)} & \colorbox{lightblue!75}{32.05} & 50.15 & \colorbox{lightblue!75}{55.56} & 51.34 & \colorbox{lightblue!75}{34.63} & \colorbox{lightblue!75}{60.43} & 14.97 & 38.10 & 9.09 & 37.70 & 24.54 & 51.16 & 16.42 & 38.68 & \colorbox{lightblue!75}{20.54} & \colorbox{lightblue!75}{48.66} \\

\hspace{2em}  + RAG\,\textit{(GT)} & 57.94 & 70.96 & 76.01 & 69.55 & \colorbox{lightblue!75}{70.74} & 84.07 & 31.29 & 57.03 & 22.99 & 51.87 & \colorbox{lightblue!75}{71.30} & 83.80 & \colorbox{lightblue!75}{51.49} & 66.42 & 35.71 & 61.61 \\

\midrule
Claude-3.5-Sonnet & \colorbox{lightblue!75}{3.06} & \colorbox{lightblue!75}{9.59} & 7.58 & 5.67 & 1.48 & 14.20 & \colorbox{lightblue!75}{0.00} & 5.78 & \colorbox{lightblue!75}{2.14} & \colorbox{lightblue!75}{9.63} & \colorbox{lightblue!75}{2.31} & 12.96 & 0.75 & \colorbox{lightblue!75}{6.72} & \colorbox{lightblue!75}{0.89} & \colorbox{lightblue!75}{10.71} \\

\hspace{2em}  + RAG\,\textit{(k=5)} & \colorbox{lightblue!75}{26.74} & \colorbox{lightblue!75}{47.55} & 44.44 & \colorbox{lightblue!75}{54.93} & 22.04 & \colorbox{lightblue!75}{54.51} & \colorbox{lightblue!75}{14.97} & \colorbox{lightblue!75}{35.94} & 5.35 & \colorbox{lightblue!75}{37.70} & 16.67 & \colorbox{lightblue!75}{47.69} & \colorbox{lightblue!75}{8.21} & \colorbox{lightblue!75}{32.09} & \colorbox{lightblue!75}{15.18} & \colorbox{lightblue!75}{43.75} \\

\hspace{2em}  + RAG\,\textit{(k=10)} & 31.49 & \colorbox{lightblue!75}{52.52} & 45.96 & \colorbox{lightblue!75}{58.81} & 28.70 & 60.25 & \colorbox{lightblue!75}{18.37} & \colorbox{lightblue!75}{44.78} & 6.42 & \colorbox{lightblue!75}{39.84} & 24.54 & \colorbox{lightblue!75}{53.94} & \colorbox{lightblue!75}{19.40} & \colorbox{lightblue!75}{41.04} & 17.86 & 44.20 \\

\hspace{2em}  + RAG\,\textit{(GT)} & \colorbox{lightblue!75}{58.19} &  \colorbox{lightblue!75}{73.87} & \colorbox{lightblue!75}{83.33} & \colorbox{lightblue!75}{70.75} & 69.81 & \colorbox{lightblue!75}{85.19} & \colorbox{lightblue!75}{32.65} & \colorbox{lightblue!75}{60.32} & 20.32 & \colorbox{lightblue!75}{56.68} & 68.98 & \colorbox{lightblue!75}{85.88} & 48.51 & \colorbox{lightblue!75}{67.91} & 35.71 & \colorbox{lightblue!75}{63.84} \\

\bottomrule
\end{tabular}
}
\caption{Performance Comparison of Different MLLMs (\%) on Chart-MRAG benchmark. The optimal retrieval configuration (\textit{SigLIP + E5-large}) is employed across all experiments to ensure controlled comparison (best scores for open-source and proprietary models highlighted in blue and red, respectively).}
\label{tab:generative_performance}
\end{table*}

%% file: acl_latex.bbl
\begin{thebibliography}{40}
\providecommand{\natexlab}[1]{#1}

\bibitem[{Abaskohi et~al.(2024)Abaskohi, Gella, Carenini, and Laradji}]{abaskohi2024fm2ds}
Amirhossein Abaskohi, Spandana Gella, Giuseppe Carenini, and Issam~H Laradji. 2024.
\newblock Fm2ds: Few-shot multimodal multihop data synthesis with knowledge distillation for question answering.
\newblock \emph{arXiv preprint arXiv:2412.07030}.

\bibitem[{Awadalla et~al.(2023)Awadalla, Gao, Gardner, Hessel, Hanafy, Zhu, Marathe, Bitton, Gadre, Sagawa et~al.}]{awadalla2023openflamingo}
Anas Awadalla, Irena Gao, Josh Gardner, Jack Hessel, Yusuf Hanafy, Wanrong Zhu, Kalyani Marathe, Yonatan Bitton, Samir Gadre, Shiori Sagawa, et~al. 2023.
\newblock Openflamingo: An open-source framework for training large autoregressive vision-language models.
\newblock \emph{arXiv preprint arXiv:2308.01390}.

\bibitem[{Chen et~al.(2024)Chen, Xiao, Zhang, Luo, Lian, and Liu}]{chen2024bge}
Jianlv Chen, Shitao Xiao, Peitian Zhang, Kun Luo, Defu Lian, and Zheng Liu. 2024.
\newblock Bge m3-embedding: Multi-lingual, multi-functionality, multi-granularity text embeddings through self-knowledge distillation.
\newblock \emph{arXiv preprint arXiv:2402.03216}.

\bibitem[{Chen et~al.(2022)Chen, Hu, Chen, Verga, and Cohen}]{chen2022murag}
Wenhu Chen, Hexiang Hu, Xi~Chen, Pat Verga, and William~W Cohen. 2022.
\newblock Murag: Multimodal retrieval-augmented generator for open question answering over images and text.
\newblock \emph{arXiv preprint arXiv:2210.02928}.

\bibitem[{Ding et~al.(2024)Ding, Ren, Huang, Luo, and Han}]{ding2024mvqa}
Yihao Ding, Kaixuan Ren, Jiabin Huang, Siwen Luo, and Soyeon~Caren Han. 2024.
\newblock Mvqa: A dataset for multimodal information retrieval in pdf-based visual question answering.
\newblock \emph{arXiv preprint arXiv:2404.12720}.

\bibitem[{Dong et~al.(2025)Dong, Chang, Goh, Li, Tang, and Liu}]{dong2025mmdocir}
Kuicai Dong, Yujing Chang, Xin~Deik Goh, Dexun Li, Ruiming Tang, and Yong Liu. 2025.
\newblock Mmdocir: Benchmarking multi-modal retrieval for long documents.
\newblock \emph{arXiv preprint arXiv:2501.08828}.

\bibitem[{Dubey et~al.(2024)Dubey, Jauhri, Pandey, Kadian, Al-Dahle, Letman, Mathur, Schelten, Yang, Fan et~al.}]{dubey2024llama}
Abhimanyu Dubey, Abhinav Jauhri, Abhinav Pandey, Abhishek Kadian, Ahmad Al-Dahle, Aiesha Letman, Akhil Mathur, Alan Schelten, Amy Yang, Angela Fan, et~al. 2024.
\newblock The llama 3 herd of models.
\newblock \emph{arXiv preprint arXiv:2407.21783}.

\bibitem[{Es et~al.(2023)Es, James, Espinosa-Anke, and Schockaert}]{es2023ragas}
Shahul Es, Jithin James, Luis Espinosa-Anke, and Steven Schockaert. 2023.
\newblock Ragas: Automated evaluation of retrieval augmented generation.
\newblock \emph{arXiv preprint arXiv:2309.15217}.

\bibitem[{Faysse et~al.(2024)Faysse, Sibille, Wu, Omrani, Viaud, Hudelot, and Colombo}]{faysse2024colpali}
Manuel Faysse, Hugues Sibille, Tony Wu, Bilel Omrani, Gautier Viaud, Céline Hudelot, and Pierre Colombo. 2024.
\newblock \href {https://arxiv.org/abs/2407.01449} {Colpali: Efficient document retrieval with vision language models}.
\newblock \emph{Preprint}, arXiv:2407.01449.

\bibitem[{Fleiss and Cohen(1973)}]{fleiss1973equivalence}
Joseph~L Fleiss and Jacob Cohen. 1973.
\newblock The equivalence of weighted kappa and the intraclass correlation coefficient as measures of reliability.
\newblock \emph{Educational and psychological measurement}, 33(3):613--619.

\bibitem[{Gao et~al.(2023)Gao, Xiong, Gao, Jia, Pan, Bi, Dai, Sun, and Wang}]{gao2023retrieval}
Yunfan Gao, Yun Xiong, Xinyu Gao, Kangxiang Jia, Jinliu Pan, Yuxi Bi, Yi~Dai, Jiawei Sun, and Haofen Wang. 2023.
\newblock Retrieval-augmented generation for large language models: A survey.
\newblock \emph{arXiv preprint arXiv:2312.10997}.

\bibitem[{Hu et~al.(2024)Hu, Gu, Dou, Fayyaz, Lu, Chang, and Peng}]{hu2024mragbench}
Wenbo Hu, Jia-Chen Gu, Zi-Yi Dou, Mohsen Fayyaz, Pan Lu, Kai-Wei Chang, and Nanyun Peng. 2024.
\newblock Mrag-bench: Vision-centric evaluation for retrieval-augmented multimodal models.
\newblock \emph{arXiv preprint arXiv:2410.08182}.

\bibitem[{Izacard et~al.(2022)Izacard, Lewis, Lomeli, Hosseini, Petroni, Schick, Dwivedi-Yu, Joulin, Riedel, and Grave}]{izacard2022few}
Gautier Izacard, Patrick Lewis, Maria Lomeli, Lucas Hosseini, Fabio Petroni, Timo Schick, Jane Dwivedi-Yu, Armand Joulin, Sebastian Riedel, and Edouard Grave. 2022.
\newblock Few-shot learning with retrieval augmented language models.
\newblock \emph{arXiv preprint arXiv:2208.03299}, 1(2):4.

\bibitem[{Koukounas et~al.(2024)Koukounas, Mastrapas, G{\"u}nther, Wang, Martens, Mohr, Sturua, Akram, Mart{\'\i}nez, Ognawala et~al.}]{koukounas2024jina}
Andreas Koukounas, Georgios Mastrapas, Michael G{\"u}nther, Bo~Wang, Scott Martens, Isabelle Mohr, Saba Sturua, Mohammad~Kalim Akram, Joan~Fontanals Mart{\'\i}nez, Saahil Ognawala, et~al. 2024.
\newblock Jina clip: Your clip model is also your text retriever.
\newblock \emph{arXiv preprint arXiv:2405.20204}.

\bibitem[{Li et~al.(2024{\natexlab{a}})Li, Wang, Xu, Wang, Feng, Kong, and Liu}]{li2024multimodal}
Lei Li, Yuqi Wang, Runxin Xu, Peiyi Wang, Xiachong Feng, Lingpeng Kong, and Qi~Liu. 2024{\natexlab{a}}.
\newblock Multimodal arxiv: A dataset for improving scientific comprehension of large vision-language models.
\newblock \emph{arXiv preprint arXiv:2403.00231}.

\bibitem[{Li et~al.(2024{\natexlab{b}})Li, Li, Wang, Jiang, Zhang, Zheng, Wang, Zheng, Yu, Huang et~al.}]{li2024benchmarking}
Yangning Li, Yinghui Li, Xingyu Wang, Yong Jiang, Zhen Zhang, Xinran Zheng, Hui Wang, Hai-Tao Zheng, Philip~S Yu, Fei Huang, et~al. 2024{\natexlab{b}}.
\newblock Benchmarking multimodal retrieval augmented generation with dynamic vqa dataset and self-adaptive planning agent.
\newblock \emph{arXiv preprint arXiv:2411.02937}.

\bibitem[{Luo et~al.(2021)Luo, Li, Wang, and Lin}]{luo2021chartocr}
Junyu Luo, Zekun Li, Jinpeng Wang, and Chin-Yew Lin. 2021.
\newblock Chartocr: Data extraction from charts images via a deep hybrid framework.
\newblock In \emph{Proceedings of the IEEE/CVF winter conference on applications of computer vision}, pages 1917--1925.

\bibitem[{Ma et~al.(2024{\natexlab{a}})Ma, Lin, Li, Chen, and Lin}]{ma-etal-2024-unifying}
Xueguang Ma, Sheng-Chieh Lin, Minghan Li, Wenhu Chen, and Jimmy Lin. 2024{\natexlab{a}}.
\newblock \href {https://doi.org/10.18653/v1/2024.emnlp-main.373} {Unifying multimodal retrieval via document screenshot embedding}.
\newblock In \emph{Proceedings of the 2024 Conference on Empirical Methods in Natural Language Processing}, pages 6492--6505, Miami, Florida, USA. Association for Computational Linguistics.

\bibitem[{Ma et~al.(2024{\natexlab{b}})Ma, Zang, Chen, Chen, Jiao, Li, Lu, Liu, Ma, Dong et~al.}]{ma2024mmlongbench}
Yubo Ma, Yuhang Zang, Liangyu Chen, Meiqi Chen, Yizhu Jiao, Xinze Li, Xinyuan Lu, Ziyu Liu, Yan Ma, Xiaoyi Dong, et~al. 2024{\natexlab{b}}.
\newblock Mmlongbench-doc: Benchmarking long-context document understanding with visualizations.
\newblock \emph{arXiv preprint arXiv:2407.01523}.

\bibitem[{Marino et~al.(2019)Marino, Rastegari, Farhadi, and Mottaghi}]{marino2019ok}
Kenneth Marino, Mohammad Rastegari, Ali Farhadi, and Roozbeh Mottaghi. 2019.
\newblock Ok-vqa: A visual question answering benchmark requiring external knowledge.
\newblock In \emph{Proceedings of the IEEE/cvf conference on computer vision and pattern recognition}, pages 3195--3204.

\bibitem[{Masry et~al.(2022)Masry, Long, Tan, Joty, and Hoque}]{masry2022chartqa}
Ahmed Masry, Do~Xuan Long, Jia~Qing Tan, Shafiq Joty, and Enamul Hoque. 2022.
\newblock Chartqa: A benchmark for question answering about charts with visual and logical reasoning.
\newblock \emph{arXiv preprint arXiv:2203.10244}.

\bibitem[{Mathew et~al.(2021)Mathew, Karatzas, and Jawahar}]{mathew2021docvqa}
Minesh Mathew, Dimosthenis Karatzas, and CV~Jawahar. 2021.
\newblock Docvqa: A dataset for vqa on document images.
\newblock In \emph{Proceedings of the IEEE/CVF winter conference on applications of computer vision}, pages 2200--2209.

\bibitem[{Methani et~al.(2020)Methani, Ganguly, Khapra, and Kumar}]{Methani_2020_WACV}
Nitesh Methani, Pritha Ganguly, Mitesh~M. Khapra, and Pratyush Kumar. 2020.
\newblock Plotqa: Reasoning over scientific plots.
\newblock In \emph{The IEEE Winter Conference on Applications of Computer Vision (WACV)}.

\bibitem[{OpenAI(2023)}]{openai2023gpt}
R~OpenAI. 2023.
\newblock Gpt-4 technical report. arxiv 2303.08774.
\newblock \emph{View in Article}, 2(5).

\bibitem[{Radford et~al.(2021)Radford, Kim, Hallacy, Ramesh, Goh, Agarwal, Sastry, Askell, Mishkin, Clark et~al.}]{radford2021learning}
Alec Radford, Jong~Wook Kim, Chris Hallacy, Aditya Ramesh, Gabriel Goh, Sandhini Agarwal, Girish Sastry, Amanda Askell, Pamela Mishkin, Jack Clark, et~al. 2021.
\newblock Learning transferable visual models from natural language supervision.
\newblock In \emph{International conference on machine learning}, pages 8748--8763. PMLR.

\bibitem[{Schwenk et~al.(2022)Schwenk, Khandelwal, Clark, Marino, and Mottaghi}]{schwenk2022okvqa}
Dustin Schwenk, Apoorv Khandelwal, Christopher Clark, Kenneth Marino, and Roozbeh Mottaghi. 2022.
\newblock A-okvqa: A benchmark for visual question answering using world knowledge.
\newblock In \emph{European conference on computer vision}, pages 146--162. Springer.

\bibitem[{Talmor et~al.(2021)Talmor, Yoran, Catav, Lahav, Wang, Asai, Ilharco, Hajishirzi, and Berant}]{talmor2021multimodalqa}
Alon Talmor, Ori Yoran, Amnon Catav, Dan Lahav, Yizhong Wang, Akari Asai, Gabriel Ilharco, Hannaneh Hajishirzi, and Jonathan Berant. 2021.
\newblock Multimodalqa: Complex question answering over text, tables and images.
\newblock \emph{arXiv preprint arXiv:2104.06039}.

\bibitem[{Team(2024)}]{sailvl}
Bytedance Douyin~Content Team. 2024.
\newblock \href {https://huggingface.co/BytedanceDouyinContent/SAIL-VL-2B/} {Sail-vl: Scalable vision language model training with high quality data curation}.

\bibitem[{Team et~al.(2024)Team, Georgiev, Lei, Burnell, Bai, Gulati, Tanzer, Vincent, Pan, Wang et~al.}]{team2024gemini}
Gemini Team, Petko Georgiev, Ving~Ian Lei, Ryan Burnell, Libin Bai, Anmol Gulati, Garrett Tanzer, Damien Vincent, Zhufeng Pan, Shibo Wang, et~al. 2024.
\newblock Gemini 1.5: Unlocking multimodal understanding across millions of tokens of context.
\newblock \emph{arXiv preprint arXiv:2403.05530}.

\bibitem[{Wang et~al.(2022)Wang, Yang, Huang, Jiao, Yang, Jiang, Majumder, and Wei}]{wang2022text}
Liang Wang, Nan Yang, Xiaolong Huang, Binxing Jiao, Linjun Yang, Daxin Jiang, Rangan Majumder, and Furu Wei. 2022.
\newblock Text embeddings by weakly-supervised contrastive pre-training.
\newblock \emph{arXiv preprint arXiv:2212.03533}.

\bibitem[{Wang et~al.(2024)Wang, Bai, Tan, Wang, Fan, Bai, Chen, Liu, Wang, Ge, Fan, Dang, Du, Ren, Men, Liu, Zhou, Zhou, and Lin}]{Qwen2VL}
Peng Wang, Shuai Bai, Sinan Tan, Shijie Wang, Zhihao Fan, Jinze Bai, Keqin Chen, Xuejing Liu, Jialin Wang, Wenbin Ge, Yang Fan, Kai Dang, Mengfei Du, Xuancheng Ren, Rui Men, Dayiheng Liu, Chang Zhou, Jingren Zhou, and Junyang Lin. 2024.
\newblock Qwen2-vl: Enhancing vision-language model's perception of the world at any resolution.
\newblock \emph{arXiv preprint arXiv:2409.12191}.

\bibitem[{Wu et~al.(2024)Wu, Jayanthi, Viswanathan, Rosenberg, Pakazad, Wu, and Neubig}]{wu2024synthetic}
Ian Wu, Sravan Jayanthi, Vijay Viswanathan, Simon Rosenberg, Sina Pakazad, Tongshuang Wu, and Graham Neubig. 2024.
\newblock Synthetic multimodal question generation.
\newblock \emph{arXiv preprint arXiv:2407.02233}.

\bibitem[{Yao et~al.(2024)Yao, Yu, Zhang, Wang, Cui, Zhu, Cai, Li, Zhao, He et~al.}]{yao2024minicpm}
Yuan Yao, Tianyu Yu, Ao~Zhang, Chongyi Wang, Junbo Cui, Hongji Zhu, Tianchi Cai, Haoyu Li, Weilin Zhao, Zhihui He, et~al. 2024.
\newblock Minicpm-v: A gpt-4v level mllm on your phone.
\newblock \emph{arXiv preprint arXiv:2408.01800}.

\bibitem[{Yu et~al.(2024)Yu, Tang, Xu, Cui, Ran, Yan, Liu, Wang, Han, Liu et~al.}]{yu2024visrag}
Shi Yu, Chaoyue Tang, Bokai Xu, Junbo Cui, Junhao Ran, Yukun Yan, Zhenghao Liu, Shuo Wang, Xu~Han, Zhiyuan Liu, et~al. 2024.
\newblock Visrag: Vision-based retrieval-augmented generation on multi-modality documents.
\newblock \emph{arXiv preprint arXiv:2410.10594}.

\bibitem[{Zhai et~al.(2023)Zhai, Mustafa, Kolesnikov, and Beyer}]{zhai2023sigmoid}
Xiaohua Zhai, Basil Mustafa, Alexander Kolesnikov, and Lucas Beyer. 2023.
\newblock Sigmoid loss for language image pre-training.
\newblock In \emph{Proceedings of the IEEE/CVF International Conference on Computer Vision}, pages 11975--11986.

\bibitem[{Zhang et~al.(2024{\natexlab{a}})Zhang, Yu, Dong, Li, Su, Chu, and Yu}]{zhang2024mm}
Duzhen Zhang, Yahan Yu, Jiahua Dong, Chenxing Li, Dan Su, Chenhui Chu, and Dong Yu. 2024{\natexlab{a}}.
\newblock Mm-llms: Recent advances in multimodal large language models.
\newblock \emph{arXiv preprint arXiv:2401.13601}.

\bibitem[{Zhang et~al.(2024{\natexlab{b}})Zhang, Patil, Jain, Shen, Zaharia, Stoica, and Gonzalez}]{zhang2024raft}
Tianjun Zhang, Shishir~G Patil, Naman Jain, Sheng Shen, Matei Zaharia, Ion Stoica, and Joseph~E Gonzalez. 2024{\natexlab{b}}.
\newblock Raft: Adapting language model to domain specific rag.
\newblock \emph{arXiv preprint arXiv:2403.10131}.

\bibitem[{Zhao et~al.(2024)Zhao, Zhang, Yu, Wang, Geng, Fu, Yang, Zhang, and Cui}]{zhao2024retrieval}
Penghao Zhao, Hailin Zhang, Qinhan Yu, Zhengren Wang, Yunteng Geng, Fangcheng Fu, Ling Yang, Wentao Zhang, and Bin Cui. 2024.
\newblock Retrieval-augmented generation for ai-generated content: A survey.
\newblock \emph{arXiv preprint arXiv:2402.19473}.

\bibitem[{Zhao et~al.(2023)Zhao, Chen, Wang, Jiao, Do, Qin, Ding, Guo, Li, Li et~al.}]{zhao2023retrieving}
Ruochen Zhao, Hailin Chen, Weishi Wang, Fangkai Jiao, Xuan~Long Do, Chengwei Qin, Bosheng Ding, Xiaobao Guo, Minzhi Li, Xingxuan Li, et~al. 2023.
\newblock Retrieving multimodal information for augmented generation: A survey.
\newblock \emph{arXiv preprint arXiv:2303.10868}.

\bibitem[{Zhou et~al.(2024)Zhou, Liu, Liu, Xiao, Wang, Zhao, Zhang, Lian, and Xiong}]{zhou2024megapairs}
Junjie Zhou, Zheng Liu, Ze~Liu, Shitao Xiao, Yueze Wang, Bo~Zhao, Chen~Jason Zhang, Defu Lian, and Yongping Xiong. 2024.
\newblock Megapairs: Massive data synthesis for universal multimodal retrieval.
\newblock \emph{arXiv preprint arXiv:2412.14475}.

\end{thebibliography}
